%% file: main.tex
\documentclass[5p,times]{elsarticle}
\usepackage[table]{xcolor}

\usepackage{graphicx}
\usepackage{amsmath}
\usepackage{amsfonts}
\usepackage{booktabs}
\usepackage{color}
\usepackage{colortbl}
\usepackage{csquotes}
\usepackage{dirtytalk}
\usepackage{dsfont}
\usepackage[T1]{fontenc}
\usepackage{float}
\usepackage{ifthen}
\usepackage{graphicx}
\usepackage{lmodern}
\usepackage{longtable}
\usepackage{natbib}
\usepackage{paralist}
\usepackage[flushleft]{threeparttable}
\usepackage{url}
\usepackage{verbatim}

\newcolumntype{L}[1]{>{\raggedright\arraybackslash}p{#1}}                                                                                                                                                              
\newcolumntype{T}[1]{>{\small\raggedright\arraybackslash}p{#1}}

\newboolean{showcomments}
\setboolean{showcomments}{true} % toggle to show or hide comments
\ifthenelse{\boolean{showcomments}}
  {\newcommand{\nb}[2]{
    \fcolorbox{gray}{yellow}{\bfseries\sffamily\scriptsize#1}
    {$\blacktriangleright$#2$\blacktriangleleft$}
   }
   
  }
  {\newcommand{\nb}[2]{}
   
  }

\newcommand{\added}[1]{\textcolor{black}{#1}}

\AtBeginDocument{%
  \providecommand\BibTeX{{%
    \normalfont B\kern-0.5em{\scshape i\kern-0.25em b}\kern-0.8em\TeX}}}

\makeatletter
\newcommand\footnoteref[1]{\protected@xdef\@thefnmark{\ref{#1}}\@footnotemark}
\makeatother

\begin{document}

\begin{frontmatter}

\title{Intelligent Software Web Agents: A Gap Analysis}

\author{Sabrina Kirrane}

\address{Institute for Information Systems and New Media, Vienna University of Economics and Business, Austria}
\ead{sabrina.kirrane@wu.ac.at}

\begin{abstract}
Semantic web technologies have shown their effectiveness, especially when it comes to knowledge representation, reasoning, and data integration. However, the original semantic web vision, whereby machine readable web data could be automatically actioned upon by intelligent software web agents, has yet to be realised.  
In order to better understand the existing technological \added{opportunities and challenges}, in this paper we examine the status quo in terms of intelligent software web agents, guided by research with respect to requirements and architectural components, coming from the agents community.
%
%We start by collating and summarising requirements and core architectural components relating to intelligent software agents. 
%
\added{We use the identified requirements to both further elaborate on the semantic web agent motivating use case scenario, and to summarise different perspectives on the requirements from the semantic web agent literature. 
We subsequently propose a hybrid semantic web agent architecture,  and use the various components and subcomponents in order to provide a focused discussion in relation to existing semantic web standards and community activities. 
Finally, we highlight open research opportunities and challenges and take a broader perspective of the research by discussing the potential for intelligent software web agents as an enabling technology for emerging domains, such as digital assistants, cloud computing, and the internet of things.}
\end{abstract}


\end{frontmatter}

%%%%%%%%%%%%%%%%%%%%%%%%%%%%%%%%%%%%%%%%%%%%%
% Introduction 
%%%%%%%%%%%%%%%%%%%%%%%%%%%%%%%%%%%%%%%%%%%%%

\input{content/iswaintroduction}

%%%%%%%%%%%%%%%%%%%%%%%%%%%%%%%%%%%%%%%%%%%%%
% agents
%%%%%%%%%%%%%%%%%%%%%%%%%%%%%%%%%%%%%%%%%%%%%

\input{content/iswaagents}

%%%%%%%%%%%%%%%%%%%%%%%%%%%%%%%%%%%%%%%%%%%%%
% Framework
%%%%%%%%%%%%%%%%%%%%%%%%%%%%%%%%%%%%%%%%%%%%%

\input{content/iswausecase}

%%%%%%%%%%%%%%%%%%%%%%%%%%%%%%%%%%%%%%%%%%%%%
% Status Quo
%%%%%%%%%%%%%%%%%%%%%%%%%%%%%%%%%%%%%%%%%%%%%

\input{content/iswafoundations}

%%%%%%%%%%%%%%%%%%%%%%%%%%%%%%%%%%%%%%%%%%%%%
% Architecture
%%%%%%%%%%%%%%%%%%%%%%%%%%%%%%%%%%%%%%%%%%%%%

\input{content/iswaarchitecture}

%%%%%%%%%%%%%%%%%%%%%%%%%%%%%%%%%%%%%%%%%%%%%
% Roadmap
%%%%%%%%%%%%%%%%%%%%%%%%%%%%%%%%%%%%%%%%%%%%%

\input{content/iswavision}

%%%%%%%%%%%%%%%%%%%%%%%%%%%%%%%%%%%%%%%%%%%%%
% Conclusion 
%%%%%%%%%%%%%%%%%%%%%%%%%%%%%%%%%%%%%%%%%%%%%

\input{content/iswaconclusion}

%%%%%%%%%%%%%%%%%%%%%%%%%%%%%%%%%%%%%%%%%%%%%
% Biblography
%%%%%%%%%%%%%%%%%%%%%%%%%%%%%%%%%%%%%%%%%%%%%

\section*{Acknowledgments}

This research is funded by the FWF Austrian Science Fund and the Internet Foundation Austria under the FWF Elise Richter and netidee SCIENCE programmes as project number V 759-N.

\bibliographystyle{abbrvnat}
\bibliography{main}
\end{document}

%% file: content/iswaintroduction.tex
% !TEX root = ../main.tex
\section{Introduction}\label{sec:intro}

At the turn of the millennium, \citet{berners2001semantic} coined the term semantic web and set a research agenda for this new research field. The authors used a fictitious scenario to describe their vision for a web of machine-readable data, which would be exploited by intelligent software agents who would carry out data centric tasks on behalf of humans. \citet{hendler2001agents} further elaborated on the intelligent software agent vision with a particular focus on the key role played by ontologies in terms of service capability advertisements that are necessary in order to facilitate interaction between autonomous intelligent software web agents. 

Several years later,  in 2007,  \citet{hendler2007all} highlighted that although the interoperability and intercommunication infrastructure necessary to support intelligent agents was available the intelligent agent vision had not yet been realised.
Almost a decade later,  in 2016,  \citet{bernstein2016new} discussed the evolution of the semantic web community and identified several open research questions, which they categorised under the following headings: (i) representation and lightweight semantics; (ii) heterogeneity, quality, and provenance; (iii) latent semantics; and (iv) high volume and velocity data.  Interestingly, the authors identified the need to better understand the needs of intelligent software web agents both from a semantics and a deployment perspective.  Hinting that the semantic web agent vision had not progressed from a practical perspective.
Further evidence that the intelligent software agents vision has received limited attention in recent years was provided by \citet{kirrane2020decade}, who used three data driven approaches in order to extract topics and trends from a corpus of semantic web venue publications from 2006 to 2015 inclusive.  The authors highlighted that the intelligent agents topic did not feature in any of the top 40 topic lists produced by the three topic and trend detection tools used for their analysis.

Nevertheless, according to \citet{luck2005agent}, a semantically rich data model, vocabularies, and ontologies, which can be used to describe media and services in a manner that facilitates discovery and composition, are key components of the proposed strategic agent technology roadmap.
Indeed, over the years, the semantic web community has produced various standards and best practices that support data integration and reasoning using web technologies \citep{bernstein2016new}.  Many of which are discussed in the recent knowledge graphs tutorial article \citep{hogan2021knowledge},  which examines the role of semantic technologies when it comes to publishing and consuming knowledge graphs.
Although several application areas (e.g., web search, commerce, social networks, finance) are discussed, there is no mention of agency.  Interestingly, agents are briefly mentioned in several places, especially in the context of Findable, Accessible, Interoperable and Reusable (FAIR) principles, however agency from a conceptual perspective is not discussed.
More broadly,  agent technology and semantic web agents in particular could potentially serve as an enabling technology for various emerging domains (e.g., digital assistants, cloud computing,  and the internet of things), especially when it comes to integration and governance.
However,  an important stepping stone to positioning intelligent software web agents as an enabling technology for more complex domains,  is to determine what standards, tools, and technologies have been proposed, and to identify open research opportunities and challenges.

Thus, motivated by the desire to better understand the status quo, we perform a focused literature review shepherded by agent requirements and architectural components commonly discussed in the literature.
Our work is guided by three primary research questions:
\begin{enumerate}
\item Which core requirements and architectural components are routinely used to guide software agent research?
\item What is the status quo in terms of intelligent software web agent research in terms of standards, tools, and technologies?
\item What are the primary opportunities and challenges for intelligent software web agents from both a requirements and an architectural perspective?
\end{enumerate}

In order to answer the aforementioned research questions we adopt an integrative literature review methodology \cite{whittemore2005integrative,torraco2005writing}.  The goal being to integrate literature,  in order to better understand the various proposals and how they relate to one another.  The objective of our analysis is not to survey all literature that could be used to realise intelligent software web agents,  but rather to use agent requirements and standard components used in agent architectures in order to perform a targeted analysis of the original intelligent software web agents motivating use case scenario and the potential solutions proposed to date.  

Towards this end, we start by examining well known literature from the agents community that relates specifically to intelligent agent requirements and architectural components. 
Next, we use the intelligent agent requirements in order to better understand the functional and non functional aspects of the envisaged intelligent software web agents, and the various perspectives on said requirements coming from the semantic web literature.
Following on from this, we use the intelligent agent architectural components in order to examine existing standards, tools, and technologies that could be used to realise the proposed hybrid agent architecture.  
We subsequently provide pointers, in the form of opportunities and challenges, that could be used to realise the semantic web agent vision.  
Finally, we discuss the potential for intelligent software web agents as an enabling technology for digital assistants, cloud computing, and the internet of things.

Our primary contributions can be summarised as follows:
\begin{inparaenum} [(i)]
\item we provide the necessary background information concerning intelligent agent requirements and architectures;
\item we introduce an agent task environment requirements assessment framework that can be used to perform a detailed analysis of various agent based use case scenarios;
\item we propose a web based hybrid agent architecture and use it to perform a gap analysis in terms of existing standards, tools, and technologies; and
\item we identify existing research opportunities and challenges, and reinforce the need for intelligent software web agents as an enabling technology for several emerging domains.
\end{inparaenum}

The remainder of the paper is structured as follows: 
\emph{Section}~\ref{sec:background} presents the necessary background information in relation to intelligent agent requirements and architectures. 
\emph{Section}~\ref{sec:swa} outlines our motivating use case scenario and presents the results of our requirements analysis. 
\emph{Section}~\ref{sec:statusquo} examines related work on intelligent software web agents from the perspective of the various agent requirements.   
\emph{Section}~\ref{sec:architecture} proposes a web based intelligent agent architecture and discusses the standards, tools,  and technologies that could be leveraged by the individual components. 
%
% \emph{Section}~\ref{sec:broaderperspective} 
\emph{Section}~6 performs a gap analysis in terms of existing standards and various research activities, summarises open research challenges and opportunities, and discusses the intelligent software agent potential beyond the original motivating use case scenario.
Finally, we present our conclusions in \emph{Section}~\ref{sec:conclusion}.

%% file: content/iswaagents.tex
% !TEX root = ../main.tex

\section{Intelligent Software Agents}\label{sec:background}

Originally robotics was the primary driver for agent based research, however the concept evolved to include software mimicking or acting on behalf of  humans (i.e., software agents) and internet robots (i.e., bots)~\cite{muller1998architectures}. In this paper we focus specifically on intelligent software agents that use web resources in order to perform data centric tasks on behalf of humans. In order to provide a theoretical grounding for our assessment of the maturity of intelligent software web agents, we provide the necessary background information on intelligent software agent requirements and provide a high level overview of the most prominent agent architectures. 
%For a detailed discussion on intelligent agent design principles the reader is referred to~\cite{russel2013artificial}. 
%
%\sabrina{Agent communication languages are out of scope (see references in~\citep{girardi2013survey}) as we can want focus more on the architecture.}

\subsection{Agent Requirements}

\citet{wooldridge1995intelligent} distinguish between weak and strong intelligent software agents. In the case of the former, the agent is capable of acting autonomously, has the ability to interact both with humans and other agents, is capable of reacting to environmental changes, and exhibits proactive goal directed behaviour. In the case of the latter, the agent exhibits each of the aforementioned traits, however these agents are conceptualised based on human like attributes, such as knowledge, belief, intention, or obligation.  In the following, we summarise different desiderata for intelligent agent behaviour and group related requirements based on the overarching function:

%\sabrina{franklin1996agent character is out of scope}

\paragraph{Basic Functions}
\begin{description}
\item[Autonomy:] Agents should manage both their state and their actions, and should be able to adapt to changes in their environment without direct intervention by humans~\citep{castelfranchi1994guarantees,wooldridge1995intelligent, maes1995artificial, franklin1996agent, jennings1996software}.
\item[Reactivity:] Agents should be able to autonomously respond to environmental changes in a timely manner~\citep{wooldridge1995intelligent, maes1995artificial, franklin1996agent, jennings1996software}.
\item[Pro-activeness:] Agents should be able to pursue proactive goal directed behaviour~\citep{wooldridge1995intelligent, maes1995artificial, franklin1996agent, jennings1996software}.
\item[Social ability:] Agents should be able to interact with humans and other agents~\citep{wooldridge1995intelligent, genesereth2012logical, franklin1996agent, jennings1996software}.
\end{description}
\paragraph{Behavioural Functions}
\begin{description}
\item[Benevolence:] Assumes that agents do not have goals that conflict with one another and thus are well meaning~\citep{rosenschein1988deals, wooldridge1995intelligent}.
\item[Rationality:] Assumes that an agent does not act in a manner that would be counter productive when it comes to achieving its goals~\citep{wooldridge1995intelligent}.
\item[Responsibility:] Involves acting according to the authority level that is entrusted to the agent either by the person or organisation that the agent represents or another agent~\citep{hermans1997intelligent}.
\item[Mobility:] Refers to the ability to move around an electronic network, for instance using remote programming in order to execute tasks on other machines~\citep{White1994TelescriptTT,wooldridge1995intelligent, franklin1996agent}.
\end{description}
\paragraph{Collaborate Functions} 
\begin{description}
\item[Interoperability, communication, and brokering services:] Agents need to be able to discover services and to interact with other agents~\citep{hermans1997intelligent}.
\item[Inter-agent coordination:] Agents need to be able to work together with other agents in order to facilitate collective problem solving~\citep{hermans1997intelligent}. 
\end{description}
\paragraph{Code of Conduct Functions}
\begin{description}
\item[Identification:] The ability to verify the identity of an agent and the person or organisation that the agent represents~\citep{hermans1997intelligent}.
\item[Security:] Involves taking measures to secure resources against accidental or intentional misuse~\citep{hermans1997intelligent}.
\item[Privacy:] Relates to being mindful of the privacy of the person or organisation that an agent represents~\citep{hermans1997intelligent}, however more broadly an agent should respect the privacy of anyone with whom it interacts.
\item[Trust:] Involves ensuring that the system does not knowingly relay false information ~\citep{wooldridge1995intelligent}.
\item[Ethics:] Involves leaving the world as it was found, limiting the consumption of scarce resources, and ensuring predictable results~\citep{hermans1997intelligent}, however in essence this could be interpreted as ensuring that agents do no harm.
\end{description}
\paragraph{Robustness Functions}
\begin{description}
\item[Stability,  performance, and scalability:] This is a broad category that relates to ensuring that agents and multi-agent systems can handle increasing workload effectively and are highly available~\citep{hermans1997intelligent, maes1995artificial, franklin1996agent}.
\item[Verification:] Relates to governance mechanisms that can be used to verify that everything works as expected~\citep{hermans1997intelligent}.
\end{description}

\subsection{Agent Architectures}

Existing architectures, encapsulating the software components and interfaces that ultimately denote an agents capabilities, can be classified as reactive, deliberative, or hybrid~\citep{wooldridge1994agent,wooldridge1995intelligent,muller1998architectures,girardi2013survey}. Reactive architectures are ideally suited for real time decision making (where time is of the essence), whereas deliberative architectures are designed to facilitate complex reasoning. Learning architectures are designed to enable the agent to improve its performance over time.   While,  hybrid architectures strive to leverage the benefits of both deliberative and reactive architectures.  In the following, we provide a high level overview of the external inputs and interfaces that are common to all architectures: 

\begin{description}
\item[Environment:] Agents act in their environment, possibly together with other agents.  When it comes to web agents, the internet is a complex space that consists of a variety of different networking technologies, devices, information sources, applications, and both human and artificial agents. 
\item[Performance Measure:]  According to~\citet{russel2013artificial}, when it comes to evaluating the effectiveness of an agent it is necessary to define success in terms of the state of the environment. Here there is a need for desirable qualities, taking into consideration that there may be conflicting goals making it necessary to assess and manage trade-offs.
\item[Sensors:] Software agents perceive the world through a variety of sensors, for instance the keyboard, cameras, microphone, network ports, etc., that can be used by agents to sense their environment. 
\item[Actuators:] Software agents are capable of performing actions via a variety of actuators, for instance the screen, printer, headphones, network ports, etc., that can be used by agents to act on their environment. 
\end{description}

\subsubsection{Reactive Architectures} 

Reactive agents are modelled on human based instinctive or reflexive behaviour~\cite{girardi2013survey}. When it comes to reactive agents there is a tight coupling between what the agent perceives and how the agent acts in the form of condition action rules~\cite{muller1998architectures}. In the following, we briefly introduce the components predominantly found in reactive architectures:

\begin{description}
\item[Condition-action rules:] The agent simply retrieves the action associated with a particular condition perceived by its sensor(s) and uses the action to give instructions to the actuator(s).
\item[State:] More advanced reactive agents maintain state in the form of information about the world, and previous interactions with the environment. Given a new perception, the agent chooses an action based on both the current perception and its history of previous perceptions. 
\end{description}

\subsubsection{Deliberative Architectures} 

Deliberative agents are rooted in the physical symbolic system hypothesis proposed by~\citet{newell2007computer}, whereby a symbolic language is used to model the environment and decisions are taken based on logical reasoning~\cite{wooldridge1994agent}. In the following, we highlight the components predominantly found in deliberative architectures:

\begin{description}
\item[Knowledge base:] The knowledge base is a symbolic encoding of both the agents knowledge of the world and the knowledge that governs its own behaviour. 
\item[Reasoning mechanism:] Logical reasoning (e.g., deduction, induction, abduction and analogy) relating to conditions perceived by the agents sensor(s), the possible alternative actions, and their impact on the environment are used to enable the agent to give instructions to its actuator(s).
\item[Goal encoding:] Goals can be used to guide the agents decision making by describing behaviours that are desirable. The reasoning mechanism is used to select the action or set of actions that will lead to the satisfaction of a given goal.
\item[Utility function:] Agents that need to choose between different possible actions or sets of actions, can be guided via a utility function that allows the agent to perform a comparative assessment,  based on preferences, such that it maximises its utility.
\end{description}

\subsubsection{Learning Architectures} 

Learning agents strive to become more effective over time,  and are deemed especially useful when the agent environment is not known a priori~\citep{russel2013artificial}.  
Although the deliberative component could potentially be enhanced with learning abilities, \citet{bryson2000cross} argues that modularity from an agent architecture perspective simplifies both design and control, thus in this paper we treat them as separate components.
Nonetheless there is a tight coupling between both the deliberative and the learning components. Generally speaking learning agent architectures are composed of four additional components, representing the performance, problem generator, critic, and learning functions:

\begin{description}
\item[Performance:] The performance component is an all encompassing term used to refer to the core inner functions of the agent.
\item[Problem generator:] The goal of the problem generator is to suggest actions that will lead to learning in the form of new knowledge and experiences.
\item[Critic:] The critic provides feedback to the agent (in the form of a reward or a penalty) with respect to its performance, which is measured against a fixed performance standard. 
\item[Learning element:] The learning element performs actions assigned by the problem generator, and uses the feedback mechanism provided by the critic to determine how the core inner functions of the agent should be amended.
\end{description}

\subsubsection{Hybrid Architectures} 

The individual reactive and deliberative architectural components described thus far can be organised into horizontal and/or vertical layers~\citep{wooldridge1994agent,wooldridge1995intelligent}. The proposed layering can be used to combat the shortcomings in terms of both reactive and deliberative architectures, however it increases the complexity of the system from a control perspective~\citep{muller1998architectures}. In the following, we briefly introduce the components predominantly found in hybrid architectures:

\begin{description}
\item[Layering:] Horizontal and/or vertical layers are used to combine different functions, such as reactivity, deliberation, cooperation, and learning.
\item[Controllers:] Controller components are necessary for planning the work, executing and monitoring activities, and managing interactions between activities. 
\end{description}

%% file: content/iswausecase.tex
% !TEX root = ../main.tex

\section{Intelligent Software Web Agents}
\label{sec:swa}

The goal of this section is to revisit the original vision for the web, whereby machine-readable data would be exploited by intelligent software agents that carry out data centric tasks on behalf of humans. We start by summarising the use case scenario originally proposed by \citet{berners2001semantic} in their seminal semantic web paper.  Following on from this, we use the task environment framework proposed by \citet{russel2013artificial} together with the requirements from the agents literature, presented in \emph{Section}~\ref{sec:background}, \added{in the form of a task environment requirements assessment,} in order to provide additional insights into the functions necessary to realise the proposed intelligent software web agents. 

\subsection{Motivating Use Case Scenario}
\label{sec:usecase}

The original semantic web vision \cite{berners2001semantic} was presented with the help of a motivating use case scenario, revolving around Pete and Lucy and their mother who has just found out that she needs to attend regular physiotherapy sessions.  Pete and Lucy ask their personal agents to prepare a physiotherapy appointment schedule such that they are able to share the chauffeuring duties. 
%http://clipart-library.com/clip-art/female-silhouette-head-9.htm
%http://clipart-library.com/clip-art/male-head-silhouette-23.htm
\bigskip

\begin{figure}[t!]
\center
  \includegraphics[width=0.8\linewidth]{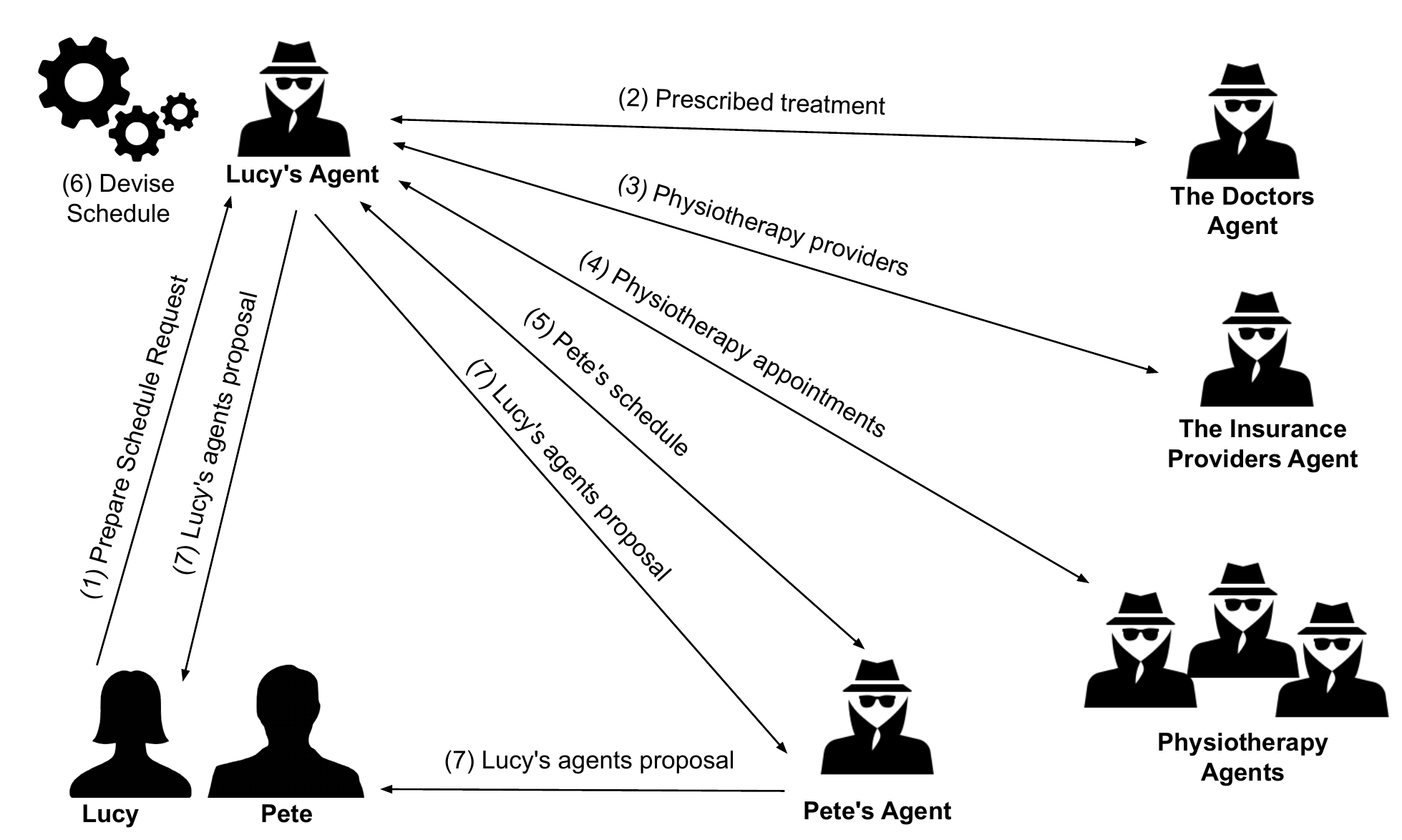}
  \caption{Physiotherapy appointment planning workflow.}
  \label{fig:lucys-agent}
\end{figure}

\noindent \paragraph{Physiotherapy appointment planning workflow}

Lucy's agent is tasked with finding a physiotherapist who: (i) is covered by their mothers insurance; (ii) has a trusted service rating of very good or excellent; (iii) is located within a 20 mile radius of their mother's home; and (iv) has appointments that work with Lucy and Pete's busy schedules. The workflow depicted in \emph{Figure}~\ref{fig:lucys-agent} can be summarised as follows: 
\begin{enumerate}[(1)]
\item Lucy requests that her agent devises a plan considering the given constraints.  
\item Lucy's agent consults with the doctor's agent in order to retrieve information relating to the prescribed treatment.
\item Lucy's agent subsequently consults with the insurance provider agent in order to find physiotherapists considering the given constraints.  
\item Lucy's agent consults the various physiotheraphy agents in order to retrieve available appointment times.
\item Lucy's agent asks Pete's agent to give her access to his schedule.
\item Lucy's agent uses the available appointment times provided by the various appointment agents, together with Pete's schedule provided by his agent, and Lucy's own schedule (that her agent already had access to) in order to prepare a physiotherapy appointment schedule considering available appointments and Pete's and Lucy's busy schedules.
\item Lucy's agent shares the proposed plan with Lucy, and Pete's agent who in turn shares it with Pete.
\end{enumerate}

\begin{figure}[t!]
\center
  \includegraphics[width=0.8\linewidth]{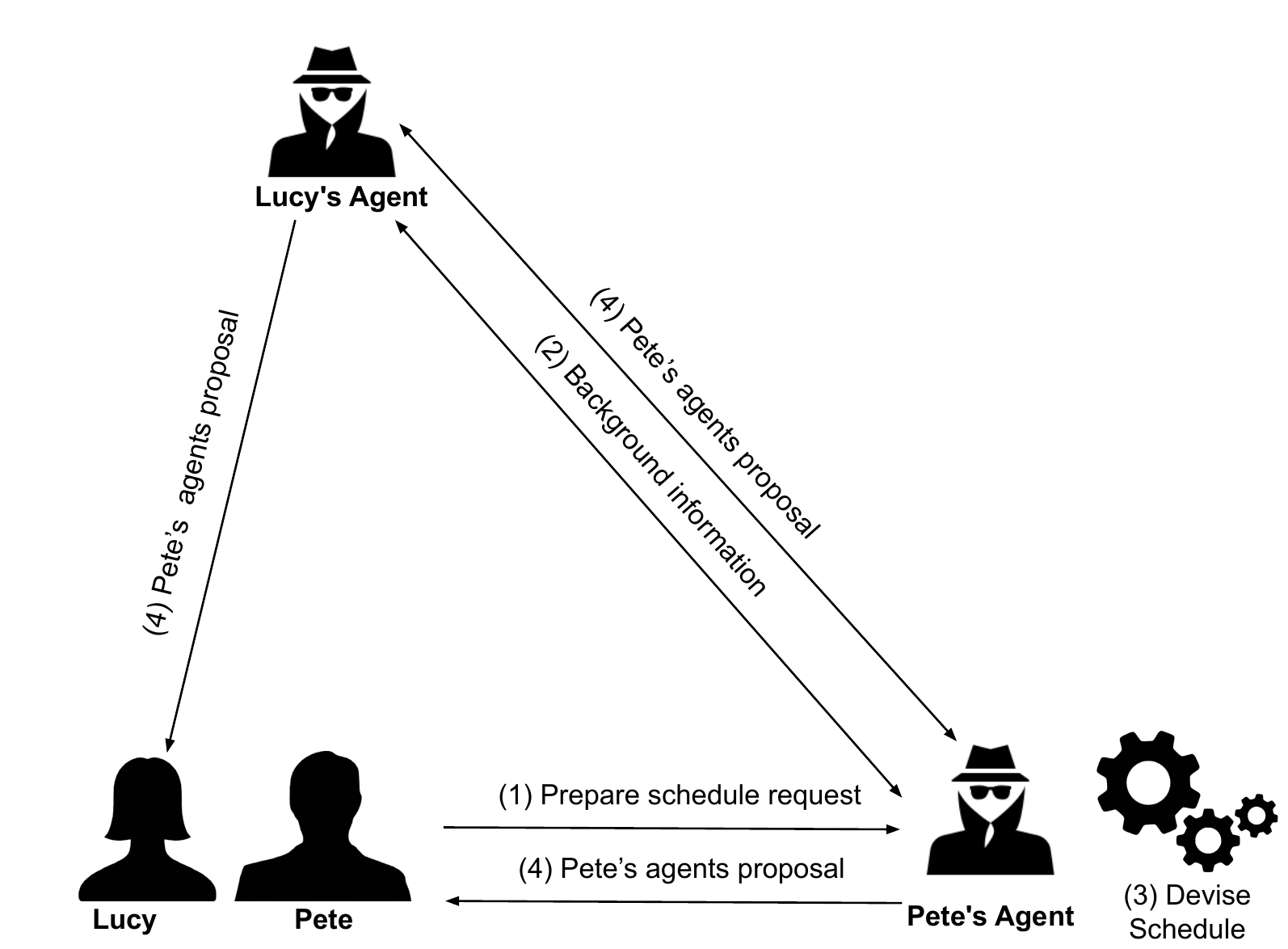}
  \caption{Physiotherapy appointment planning workflow with additional constraints.}
  \label{fig:petes-agent}
\end{figure}

\noindent \paragraph{Physiotherapy appointment planning workflow with additional constraints}

Given the physiotherapist is quite far from Pete's work and the appointment times coincide with the lunch time rush hour, Pete instructs his agent to redo the task with stricter constraints on location and time.  The workflow depicted in \emph{Figure}~\ref{fig:petes-agent} is described as follows: 
\begin{enumerate}[(1)]
\item Pete requests that his agent propose a new plan that takes into consideration the additional location and time constraints. 
\item Pete's agent obtains all relevant background information relating to the initial proposal from Lucy's agent.
\item Pete's agent uses the new constraints, together with the available appointment times and information related to Pete's and Lucy's schedules, in order to prepare a new plan, with two compromises: (i) Pete needs to reschedule some conflicting appointments, and (ii) the provider was not on the insurance companies list, thus his agent verified that the service provider was eligible for reimbursement using an alternative mechanism.
\item Pete's agent shares the proposed plan and the compromises with Pete, and Lucy's agent, who in turn shares it with Lucy.
\end{enumerate}

\begin{figure}[t!]
\center
  \includegraphics[width=0.8\linewidth]{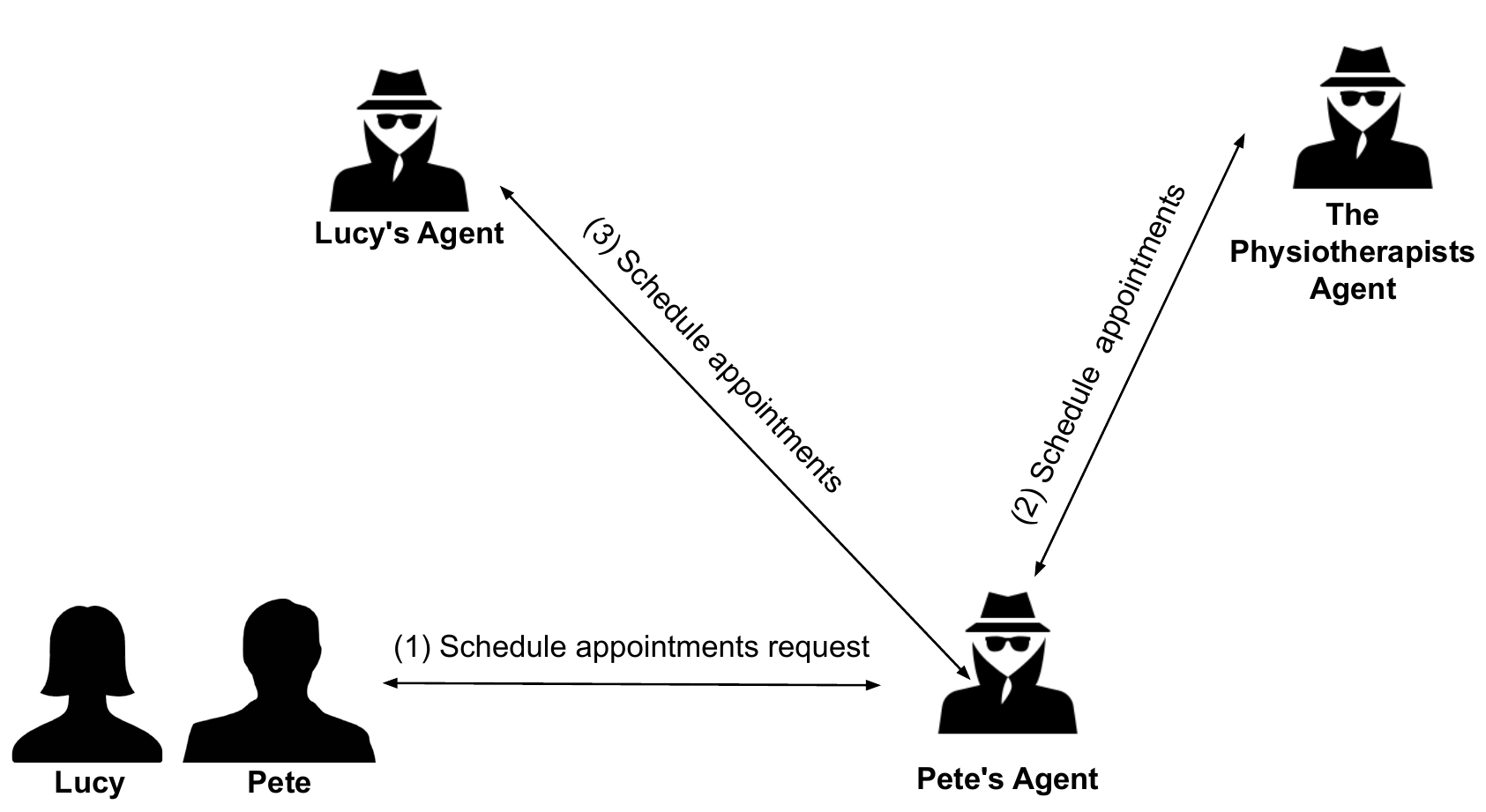}
  \caption{Physiotherapy appointment conformation workflow.}
  \label{fig:petes-agent-booking}
\end{figure}

\noindent \paragraph{Physiotherapy appointment conformation workflow}

Both Pete and Lucy agree to the plan, and Pete instructs his agent to make the appointments with the physiotherapist and update their schedules accordingly. The workflow depicted in \emph{Figure}~\ref{fig:petes-agent-booking} is outlined below: 
\begin{enumerate}[(1)]
\item Pete instructs his agent to confirm the appointments and update his schedule accordingly.
\item Pete's agent makes the booking with the physiotherapists agent. 
\item Pete's agent instructs Lucy's agent to update her schedule. 
\end{enumerate}

\bigskip

%\noindent Later Pete asks his agent to explain who it was able to verify the service provider was eligible for reimbursement.  \sabrina{For completeness I should either add a workflow or remove this.}

\subsection{A Task Environment Assessment}

Next we use the Performance Measure,  Environment,  Actuators,  and Sensors (PEAS) assessment criteria proposed by~\citet{russel2013artificial} to get a better understanding of high level goals and to examine the external interfaces and inputs.  

\subsubsection{Information Agents}

In the given scenario, the doctor's agent and the insurance company service provider's agent can be classified as \emph{information agents}. Given a request the agent uses the search parameters submitted with the request together with its own knowledge base in order to return an appropriate response.  The details of our PEAS assessment can be found below: 

{\raggedright
\begin{description}
\item[Agent:] The doctor's agent.
\item[Performance Measure:] The treatment information is provided.
\item[Environment:] The patient's agent. 
\item[Sensors:] The doctor's agent \texttt{RetrievePrescribedTreatment} interface. 
\item[Actuators:] The doctor's agent \texttt{RetrievePrescribedTreatment} interface.
\end{description}
}

\bigskip

{\raggedright
\begin{description}
\item[Agent:] The insurance company service provider's agent. 
\item[Performance Measure:] The physiotherapy service provider information is provided.
\item[Environment:] The client's agent. 
\item[Sensors:] The insurance company service provider's agent \texttt{RetrieveServiceProviderInfo} interface.
\item[Actuators:] The insurance company service provider's agent \texttt{RetrieveServiceProviderInfo} interface.
\end{description}
}

\subsubsection{Booking Agents}

The physiotherapy appointment agent has two functions: (i) to provide information about available appointments; and (ii) to accept and confirm appointment requests.  Thus this agent can be classified as a \emph{booking agent} that allows for appointments to be scheduled based on availability.  The details of our PEAS assessment is as follows:

{\raggedright
\begin{description}
\item[Agent:] The physiotherapy provider agents.
\item[Performance Measure:] The available appointments are provided,  and/or the requested appointments are confirmed.
\item[Environment:] The client's agent. 
\item[Sensors:] The physiotherapy appointment agent \texttt{RetrieveAvailableAppointments} and \texttt{BookAvailableAppointments} interfaces.
\item[Actuators:] The physiotherapy appointment agent \texttt{RetrieveAvailableAppointments} and \texttt{BookAvailableAppointments} interfaces.
\end{description}
}

\subsubsection{Personal Planning Agents}

Taking the given scenario into consideration, both Pete and Lucy's personal agents can be classified as \texttt{planning agents} that are tasked with providing an optimal plan based on the given constraints in terms of available treatments, location, time, etc.  In addition, the agents need to work together in order to find a schedule that works for both Pete and Lucy.

{\raggedright
\begin{description}
\item[Agent:] Lucy's personal agent.
\item[Performance Measure:] The physiotherapist is covered by insurance,  the physiotherapist is located near their mothers house,  and the appointments fit with Pete and Lucy's schedules. 
\item[Environment:] The doctor's agent,  the insurance company agent,  the physiotherapy provider agents, Pete's agent, and Lucy.
\item[Sensors:] Personal agent \texttt{PrepareSchedule}, \texttt{RequestInfo}, \texttt{MakeBooking} interfaces,  and a web user interface.
\item[Actuators:] Personal agent \texttt{PrepareSchedule}, \texttt{RequestInfo}, \texttt{MakeBooking} interfaces,  and a web user interface.
\end{description}
}

\bigskip

{\raggedright
\begin{description}
\item[Agent:] Pete's personal agent.
\item[Performance Measure:] The physiotherapist is covered by insurance,  the physiotherapist is located near their mothers house,  appointments fit with Pete and Lucy's schedules,  the physiotherapist is near Pete's work, and appointments are not during busy traffic periods.
\item[Environment:] (Potential) The doctor's agent,  the insurance company agent,  the physiotherapy provider agents,  Lucy's agent, and Pete.
\item[Sensors:] Personal agent \texttt{PrepareSchedule}, \texttt{RequestInfo}, \texttt{MakeBooking} interfaces,  and a web user interface.
\item[Actuators:] Personal agent \texttt{PrepareSchedule}, \texttt{RequestInfo}, \texttt{MakeBooking} interfaces,  and a web user interface.
\end{description}
}

\subsection{A Task Environment Requirements Assessment}
\label{sec:framework}

Unfortunately, the PEAS assessment does not provide any guidance with respect to the inner workings of the various agents, even though such information is necessary in order to determine the various architectural components and how they interact.  Thus, in this section we propose a task environment requirements assessment and use it to perform a more detailed analysis of our motivating use case scenario.

\begin{table*}[!t]
\scriptsize
\rowcolors{2}{gray!25}{white}
\caption{Basic functional requirements assessment.}
\begin{tabular}{p{3cm} p{4.5cm} p{4.5cm} p{4.5cm}}
\toprule
\rowcolor{gray!50}
&  \textbf{Information Agent} & \textbf{Booking Agent} & \textbf{Scheduling Agent}  \\
\textbf{Autonomy}  & handles information requests & handles information and booking requests & consult relevant sources,  devise an optimal schedule \\ 
\textbf{Reactivity}  & immediate response & immediate response & immediate response where possible \\   
\textbf{Pro-activeness} & information goal & booking goal & scheduling goal,  explore alternatives  \\      
\textbf{Social ability}  & humans and agents & humans and agents & humans and agents \\      
\bottomrule                
\end{tabular}
\label{tab:BasicAssessment}
\end{table*}

\bigskip
\noindent We start by examining the \emph{basic functions}, summarised in \emph{Table}~\ref{tab:BasicAssessment}, that are needed to determine the type of architecture (i.e., reactive, deliberative, learning, and hybrid) required:
\begin{description}
\item[Autonomy:] All three agent types are able to perform tasks without human interaction.    
\item[Reactivity:] The information and booking agents are simple request response agents, however scheduling agents need to both interact with other agents and to examine possible solutions to the task that they have been given.
\item[Pro-activeness:] Although all three agents exhibit goal directed behaviour, scheduling agents would be classified as more pro-active as they need to explore various alternatives.
\item[Social ability:] When it comes to the information and booking agents, although the scenario focuses primarily on agent to agent interaction, all three agent types need to be able to interact with humans and agents.   
\end{description}

\begin{table*}[!t]
\scriptsize
\rowcolors{2}{gray!25}{white}
\caption{Behavioural functional requirements assessment.}
\begin{tabular}{p{3cm} p{4.5cm} p{4.5cm} p{4.5cm}}
\toprule
\rowcolor{gray!50}
&  \textbf{Information Agent} & \textbf{Booking Agent} & \textbf{Scheduling Agent}  \\
\textbf{Benevolence}  & well meaning by design & well meaning by design & well meaning by design,  manages conflicting goals \\                                  
\textbf{Rationality}  & rational by design  & rational by design & rational by design  \\ 
\textbf{Responsibility}  & provides access to information  & provides access to information,  complete the booking & manages access to information,  finds an optimal schedule given a set of constraints.  \\  
\textbf{Mobility}  & - & - & interacts with several other agents \\     
\bottomrule    
\end{tabular}
\label{tab:BehaviouralAssessment}
\end{table*}

\bigskip
\noindent Next, we examine the various \emph{behavioural functions}, summarised in \emph{Table}~\ref{tab:BehaviouralAssessment}, that govern how our agents are expected to act:
\begin{description}
\item[Benevolence:] We assume that information, booking, and scheduling agents are well meaning, however it is conceivable that different personal agents in the broader sense may have conflicting goals. 
\item[Rationality:] Agents should be designed in order to ensure that agents do not act in a manner that would be counter productive when it comes to achieving their goals.
\item[Responsibility:] In the case of all three agent types, there is a tight coupling between responsibility and the overarching goals of the various agents (namely, providing access to the requested information, completing the booking,  and finding an optimal schedule given a set of constraints).
\item[Mobility:] In the given use case scenario, the agents either request information from other agents or request that other agents perform a specific action, thus we assume that the agents are immobile. 
\end{description}

\begin{table*}[!t]
\scriptsize
\rowcolors{2}{gray!25}{white}
\caption{Collaborative functional requirements assessment.}
\begin{tabular}{p{3cm} p{4.5cm} p{4.5cm} p{4.5cm}}
\toprule
\rowcolor{gray!50}
&  \textbf{Information Agent} & \textbf{Booking Agent} & \textbf{Scheduling Agent}  \\
\textbf{Interoperability} & agreed/common schema  & agreed/common schema & agreed/common schema\\
\textbf{Communication} & pull requests & pull requests & push and pull requests \\
\textbf{Brokering services} & collects information from multiple service providers & handles bookings for multiple service providers and clients & collects information from a variety of  sources \\
\textbf{Inter-agent coordination}  & - & - & agents support each other via information sharing\\ 
\bottomrule    
\end{tabular}
\label{tab:CollaborativeAssessment}
\end{table*}

\bigskip
\noindent Next we examine the \emph{collaborate functions}, summarised in \emph{Table}~\ref{tab:CollaborativeAssessment}, that can be used to determine how agents interact with humans and other agents, and how internal communication between agent components should work:

\begin{description}
\item[Interoperability:] The requester needs to know which services it can call and how it can process the responses, thus there is a need for a schema that is understood by both the requester and the requestee.
\item[Communication:] In the given scenario, we assume that agents cater for pull requests via services, and in the case of scheduling agents push notifications on completion of a task.
\item[Brokering services:] Information agents are responsible for gathering information from multiple providers, the booking agents need to handle bookings for multiple service providers, and the personal agents are tasked with collecting the information needed in order to complete its task.
\item[Inter-agent coordination:] In the given scenario, the personal agents engage in a form of collaborative problem solving.  The personal agents do not work collectively but rather support each other via information sharing (e.g.,  Pete's agent obtained all relevant background information relating to its task from Lucy's agent).
\end{description} 
 
\begin{table*}[!t]
\scriptsize
\rowcolors{2}{gray!25}{white}
\caption{Code of conduct functional requirements assessment.}
\begin{tabular}{p{3cm} p{4.5cm} p{4.5cm} p{4.5cm}}
\toprule
\rowcolor{gray!50}
&  \textbf{Information Agent} & \textbf{Booking Agent} & \textbf{Scheduling Agent}  \\
\textbf{Identification} &  may need to differentiate between public and private information providers & may need to differentiate between public and private information providers, some service providers may work on an honours system & needs to differentiate between public and private information, may need to prove who they represent\\ 
\textbf{Security}  & protect against unauthorised access,  inappropriate use,  and denial of service & protect against unauthorised access,  inappropriate use,  and denial of service  & protect against unauthorised access,  inappropriate use,  and denial of service \\
\textbf{Privacy} &  may need to handle personal information & needs to handle personal information & handles personal information appropriately \\     
\textbf{Trust}  & manages information accuracy & manages information accuracy &  manages information and scheduling accuracy,  reliable and fair,  provides transparency and explainability,  robust in terms of verification and validation   \\      
\textbf{Ethics}  & do no harm by design & do no harm by design &  do no harm by design\\ 
\bottomrule    
\end{tabular}
\label{tab:CodeOfconductAssessment}
\end{table*}

\bigskip
\noindent The \emph{code of conduct functions}, summarised in \emph{Table}~\ref{tab:CodeOfconductAssessment}, refer to the security, privacy,  and ethical requirements that need to be built into the system: 
\begin{description}
\item[Identification:] In the case of the information agent it may or may not be necessary to authenticate the requester, for instance the personal heath records are available only to patients or their agents, however service provider information is usually public.  Similarly, given that some service providers work on an honours system, authentication may or may not be needed in order to make a booking, however in both cases the client would need to be identifiable.  When it comes to the scheduling agents, considering the amount of personal data needed by the agents in our motivating scenario, they would need to be able to authenticate their owners, and also the other personal agents with whom they interact.
\item[Security:] The system should be designed to protect against unauthorised access to, and inappropriate use of, data, as well as protecting against denial of service attacks. 
\item[Privacy:] In the case of the information agents, it is necessary to differentiate between public and private information providers. However, in all other cases,  agents will need to manage personal data and thus they need to adhere to the respective data protection legislation.
\item[Trust:] All three agent types need to be able to assess if they can trust the providers that they interact with,  and ideally should be able to assess if the information they obtain from others is indeed correct.  In this context trust is a broad concept linked to reliability, fairness, transparency,  explainability,  verification and validation.      
\item[Ethics:] Although there are many things that could be discussed in detail under the ethics umbrella, here we envisage systems that do no harm, thus the agents should avoid behaving in a way that would bring about negative consequence either for the agent itself, or the agents and humans it interacts with.
\end{description} 
 
\begin{table*}[!t]
\scriptsize
\rowcolors{2}{gray!25}{white}
\caption{Robustness functional requirements assessment.}
\begin{tabular}{p{3cm} p{4.5cm} p{4.5cm} p{4.5cm}}
\toprule
\rowcolor{gray!50}
&  \textbf{Information Agent} & \textbf{Booking Agent} & \textbf{Scheduling Agent}  \\
\textbf{Stability} & available reliable \& secure &  available reliable \& secure & available reliable \& secure\\           
\textbf{Performance} & provides real time access to information & provides real time access to information & provides real time access to information, timely goal completion \\           
\textbf{Scalability} & handles increasing \newline requests \& data &  handles increasing \newline requests \& data & handles increasing \newline requests,  data, \& task \newline complexity\\           
\textbf{Verification}  & checks information is \newline correct &  checks information is \newline correct & checks  information is \newline correct, the reasoning is \newline explainable \\           
\bottomrule    
\end{tabular}
\label{tab:PerformanceAssessment}
\end{table*}

\bigskip
\noindent Finally, the \emph{robustness} requirements assessment, summarised in \emph{Table}~\ref{tab:PerformanceAssessment}, defines criteria that should be used to determine the effectiveness of the architecture in terms of both functional and non functional requirements:

\begin{description}
\item[Stability:] Stability is an all encompassing term used for availability, reliability, and security. Such metrics have an important role to play when it comes to evaluating the effectiveness of any system. 
\item[Performance:] While information and booking agents need to be able to respond in real time, the scheduling agents will require time in order to source and analyse the data needed to derive an optimal schedule that satisfies a given set of constraints.
\item[Scalability:] All three agent types need to be able to scale with increasing data and increasing requests. 
\item[Verification:] All three agent types need  mechanisms that can be used to verify that everything works as expected.  In addition, scheduling agents need to be able to explain what information sources they used and the logic behind their proposal.  
\end{description} 

%% file: content/iswafoundations.tex
% !TEX root = ../main.tex

\section{Intelligent Software Web Agents: Requirements}\label{sec:statusquo}

In this section, we take a closer look at intelligent software agents and how the various requirements are perceived from a semantic web perspective.  Considering the broad nature of the topic, the goal is not to summarise all relevant literature, but rather to better understand the different perspectives on the various requirements introduced in \emph{Section}~\ref{sec:background} and discussed specifically in the context of the original semantic web agent use case scenario in \emph{Section}~\ref{sec:swa}. 

%The overarching goal of our integrative literature review \cite{whittemore2005integrative},  is to 

\begin{table*}[!t]
\scriptsize
\rowcolors{2}{gray!25}{white}
\caption{Intelligent software web agents basic function perspectives.}
\begin{tabular} 
{p{4cm} p{3cm} p{3cm} p{3cm} p{3cm}}
\toprule
%\rowcolor{gray!50}
&
\textbf{Autonomy} &
\textbf{Reactivity} & 
\textbf{Pro-activeness} & 
\textbf{Social ability} 
 \\
\bottomrule  
\citet{artz2007survey} & autonomy \& trust & - & - & social  networks \&  reputation  \\
\citet{boley2001design} & - & event condition action rules & - & - \\
\citet{bonatti2010reactive} &  - & trust negotiation between agents & - & social network use case  \\
\citet{bryson2002toward}  \newline  \citet{bryson2003agent} & semi-autonomous modules \&  autonomous  agents & reactive plans & - & - \\
\citet{buhler2005towards} & autonomous agents \& workflows & - & semantic web  services \& behavioural descriptions & social  structures \&  workflows \\
\citet{buoncompagni2017ros} & learning agents &  - &  learning ability & - \\
\citet{challenger2018development} & - & - &  belief-desire intention & - \\
\citet{chiu2005towards} & autonomous agents & - & believe-desire-intention framework \& ontologies & - \\
\citet{demarchi2018integration} & - & - &  belief-desire intention & - \\
\citet{dong2015bowl} & - & - & belief-desire intention & - \\
\citet{ermolayev2004towards} & autonomous agents & - & collaborative goals & social commitments  \& conventions \\
\citet{fornara2018using}  \newline  \citet{fornara2019using} &  policy agents & -&  - & -  \\
\citet{garcia2008combining}  \newline  \citet{garcia2009ontology} &  autonomous agents & - & semantic web services \& behavioural descriptions & -  \\
\citet{ghanadbashi2021using} & learning agents & - & learning ability & - \\
\citet{gomes2015procedure} &  -  &  event condition transaction language & - & - \\
\citet{harth2017specifying} & - & condition action rule language & - & - \\
\citet{huhns2002agents} & autonomy \& co-operation  & - & - & -  \\
\citet{jochum2019data} & -  & event condition action rules &  - & - \\
\citet{kafer2018rule} & - & condition action rule language & - & - \\
\citet{khalili2008framework} & autonomous agents & environmental changes & goals & communication language  \\
\citet{kootbally2017overview} & learning agents & - &  learning ability & - \\
\citet{ksystra2016formal} & - & event condition action rules & - & - \\
\citet{leite2013using} \newline \citet{leite2014case} & learning agents & - &  learning ability & - \\
\citet{merkle2019cooperative} & learning agents &  - &  learning ability & - \\
\citet{paolucci2003autonomous} & autonomous semantic web services & - & - & -  \\
\citet{papamarkos2003event} \newline \citet{papamarkos2004rdftl} & - & event-condition-action rule language & - & -  \\
\citet{payne2008web} & learning agents & environmental changes & goals & social awareness \\
\citet{pham2011practical} &-& -&  goals & - \\
\citet{poulovassilis2006event} & - & event condition action rules & - & - \\
\citet{rajpathak2004ontological} &-& -& goals & - \\
\citet{sycara2004dynamic} & discover \&  synchronisation & - & - & -\\
\citet{tamma2008semantic} &  service providers \&   autonomous agents & - & - & - \\
\citet{tamma2004serse} & autonomous system components & update indexes  & update indexes  & collaborative query answering  \\
\citet{terziyan2008smartresource} & autonomous resources & - & modelling context,  dynamics,  \&  coordination & -  \\
\citet{tonti2003semantic} & policy agents & - & - & social awareness \\
\citet{van2015semantic} & normative agents & - & -  & socially adaptive agents \\
\bottomrule   
\end{tabular}
\normalsize
\label{tab:sq-basic}
\end{table*}

\subsection{Basic Functions}

The works categorised in \emph{Table}~\ref{tab:sq-basic} and discussed below describe how intelligent agents could leverage semantic technologies, usually from a theoretical perspective.

\paragraph{Autonomy}
\citet{paolucci2003autonomous} focus on service provision and usage, using the term autonomous semantic web services to refer to services that are capable of reconfiguring their interaction patterns, such that it is possible for them to react to changes with minimal human involvement.  
Several authors~\cite{chiu2005towards, buhler2005towards, ermolayev2004towards, khalili2008framework, terziyan2008smartresource, garcia2008combining, garcia2009ontology} focus specifically on autonomous agents and how they can leverage web services. 
While, \citet{bryson2002toward,bryson2003agent} take a more conservative view referring to agent behaviours as semi-autonomous intelligent modules.  \citet{artz2007survey} in turn discuss the relationship between autonomy and trust. 
While, \citet{van2015semantic} focus on adaptability, discussing how agents can adapt their behaviour in order to comply with norms. 
\citet{tamma2004serse} identify autonomous components as a desiderata for searching the semantic web and \citet{sycara2004dynamic} highlight the key role played by brokers when it comes to discovery and synchronisation between autonomous agents. More generally,\citet{tamma2008semantic} argue that the sheer scale and heterogeneity of knowledge and services available on the web calls for autonomy not only on the part of the data and service providers but also the intelligent agents that are best placed to adapt to such dynamic, uncertain, and large scare environments. 
\citet{payne2008web,leite2013using, leite2014case, buoncompagni2017ros, kootbally2017overview, merkle2019cooperative} and \citet{ghanadbashi2021using} discuss autonomy from a learning perspective, highlighting the need for agents to be self-aware by building up a knowledge base that allows them to learn alternative strategies and solutions that can be used to fulfil future goals.  
While, \citet{huhns2002agents} focuses specifically on the tension between autonomy and co-ordination when it comes to inter-agent co-operation.  In particular, the author highlights the need for extending web service standards to cater for federated servers and co-operating clients.
Whereas,  the autonomous agent architectures proposed by \citet{fornara2018using, fornara2019using, tonti2003semantic} and \citet{van2015semantic} are designed to cater for constraints in the form of policies or norms.

\paragraph{Reactivity}

\citet{bryson2002toward,bryson2003agent} discuss how an agent-oriented approach to software engineering, entitled behaviour-oriented design, can be used to define reactive intelligent software web agents that are capable of managing interconnected (possibly conflicting) reactive plans.
\citet{boley2001design, papamarkos2003event, papamarkos2003event, papamarkos2004rdftl,poulovassilis2006event,gomes2015procedure, ksystra2016formal} and \citet{jochum2019data} propose solutions that can be used to encode reactive functionality in the form of event-condition-action rules. Whereas, the architecture proposed by \citet{kafer2018rule} makes use of simple condition-action rules.
The discussion on web services from an agents perspective by \citet{payne2008web} and the framework proposed by \citet{khalili2008framework} consider reactivity in terms of an agents response to environmental changes.
\citet{bonatti2010reactive} in turn propose a formal framework that can be use to express and enforce reactive policies, while at the same time catering for trust negotiation between agents.
While, \citet{tamma2004serse} discuss the role played by both reactive and pro-active components in their searching for semantic web content system,  where a reactive approach is used to keep indexes up to date.

\paragraph{Pro-activeness}

\citet{buhler2005towards} argue that semantic web services together with semantic behavioural description can be used by agents in order to achieve pro-active behaviour.  The proposed approach also serves as a foundation for the ontology based intelligent agent framework proposed by~\citet{garcia2009ontology, garcia2008combining}.
\citet{rajpathak2004ontological, khalili2008framework, payne2008web} and \citet{pham2011practical} consider pro-activeness in terms of goal directed agent behaviours,  with \citet{ermolayev2004towards} also considering pro-activeness in terms of collaborative goals in a multi-agent system.
The agents proposed by \citet{chiu2005towards, dong2015bowl, demarchi2018integration, challenger2018development} all employ the belief–desire–intention software model.
While, \citet{payne2008web,leite2013using, leite2014case, buoncompagni2017ros, kootbally2017overview, merkle2019cooperative} and \citet{ghanadbashi2021using} examine how agents can be enhanced with pro-active learning ability.
In turn, the multi-agent information system infrastructure proposed by~\citet{chiu2005towards} is rooted in the believe-desire-intention framework whereby ontologies are used to encode knowledge that the agent acts upon.
While,  \citet{terziyan2008smartresource} argues that semantic web standards need to be extended in order to cater for context,  dynamics,  and co-ordination,  necessary to facilitate proactivity between agents. 
In the context of their searching for semantic web content system,  according to \citet{tamma2004serse} a proactive approach should be used by agents to inform other agents of any local changes.

\paragraph{Social ability}

\citet{tamma2004serse} are guided by requirements relating to searching the semantic web whereby agents collaborate in order to answer queries. \citet{artz2007survey} highlight the importance of social networks when it comes to trust in and among agents. 
\citet{buhler2005towards} discuss the role of agent cooperation and coordination from a workflow enactment perspective.    While, \citet{ermolayev2004towards} identify the need for social commitments and conventions to regulate group activities.
Both \citet{khalili2008framework} and \citet{payne2008web} highlight the fact that agents are socially aware,  however \citet{khalili2008framework} furthers the notion by highlighting the importance of a common agent communication language.
\citet{bonatti2010reactive} demonstrate the effectiveness of their reactive policies and negotiation framework using a social network communication tool.
The agents proposed by \citet{van2015semantic} are socially adaptive agents in the sense that they strive towards norm compliance.  \citet{tonti2003semantic} also adopt a social perspective, highlighting the need for policies that can constrain agent behaviour.

\begin{table*}[!t]
\scriptsize
\rowcolors{2}{gray!25}{white}
\caption{Intelligent software web agents behavioural function perspectives.}
\begin{tabular} 
{p{4cm} p{3cm} p{3cm} p{3cm} p{3cm}}
\toprule
%\rowcolor{gray!50}
&
\textbf{Benevolence} &
\textbf{Rationality} & 
\textbf{Responsibility} & 
\textbf{Mobility} 
 \\
\bottomrule  
\citet{artz2007survey} & benevolence \& trust & - & - & - \\
\citet{bryson2003agent} & - & - & data retention & - \\
\citet{chen2004intelligent} & - & - & context  broker agent & service discovery \\
\citet{demarchi2018integration} & - & - &  agent platform components & - \\
\citet{ermolayev2004towards} & self-interest \& benevolence.  & rationality \& group dynamics & - & service  mobility \& availability \\
\citet{gandon2019web} & benevolence \&  societal benefit & - & - & -\\
\citet{garcia2008combining}  \citet{garcia2009ontology} & - & - & agent types, roles, \& responsibilities & - \\
\citet{jutla2006pecan} & benevolence,  trust  \& integrity & - & - & - \\
\citet{khalili2008framework} & conflicting goals & goal oriented decision making & - & move around a \newline network \\
\citet{paolucci2003autonomous}   & - & - & web service architecture & - \\
\citet{payne2008web} & - & goal oriented decision making & - & - \\
\citet{scioscia2014mobile}  & - & - & - & service discovery \\
\citet{sheshagiri2004using}  & - & - & - & service discovery \\
\citet{tamma2008semantic} & - & partial \& updated  knowledge & - & - \\
\bottomrule   
\end{tabular}
\normalsize
\label{tab:sq-behaviour}
\end{table*}

\subsection{Behavioural Functions}

The works presented in \emph{Table}~\ref{tab:sq-behaviour} and discussed in more detail below provide different perspectives on the behavioural functions that could potentially be built into intelligent software web agents.

\paragraph{Benevolence}

Both \citet{artz2007survey} and \citet{jutla2006pecan} briefly mention benevolence in the context of making decisions,  however \citet{artz2007survey} qualifying its use as a willingness to expend the effort needed to establish trust.  
\citet{khalili2008framework} focus on the assumption that benevolent agents do not have conflicting goals.
\citet{ermolayev2004towards} discuss benevolence from a multi-agent group utility perspective, highlighting the need to balance self-interest and benevolence.  
While, \citet{gandon2019web} focus on the societal benefit of the web, arguing that artificial intelligence based applications need to be benevolent by design.

\paragraph{Rationality}

According to \citet{payne2008web} agents need to act rationally when it comes to decision making, for instance by considering the utility gain in terms of a reward or a perceived advantage.  In addition to defining rational behaviour,  \citet{khalili2008framework} also specifically state that it is assumed that agents don't act in a counter productive manner.  While, \citet{ermolayev2004towards} discuss rationality from a multi-agent perspective, focusing on the need to balance individual-rationality from a self-interest perspective and benevolence when it comes to group dynamics.
Whereas, \citet{tamma2008semantic} identify the need for bounded rational deliberation when it comes to partial knowledge and updates to existing knowledge.

\paragraph{Responsibility}

\citet{paolucci2003autonomous} discuss responsibility purely from a web service architecture perspective.  While, \citet{bryson2003agent} focus on responsibility from a data retention perspective.  The context broker architecture proposed by \citet{chen2004intelligent} focuses specifically on the responsibilities of the context broker agent which is at the core of the proposed meeting system.  \citet{demarchi2018integration} in turn examine responsibility from an architectural perspective., identifying the need for responsible components.  While,  \citet{garcia2008combining, garcia2009ontology} take an intelligent agent perspective, identifying several different types of agents that are differentiated from one another via roles and responsibilities.

\paragraph{Mobility}

\citet{khalili2008framework} list mobility (in terms of ability to move around a network) as one of the requirements of an agent based system.
\citet{ermolayev2004towards} highlight the need for mobile agents in order to ensure the robustness of the system from an availability and a performance perspective.  Several authors \cite{sheshagiri2004using, chen2004intelligent, scioscia2014mobile} highlight the key role played by semantic web services when it comes to service discovery in mobile and ubiquitous environments.  While, \citet{outtagarts2009mobile} performs a broad survey of mobile agent applications, with semantic web services being one of them.

\begin{table*}[!t]
\scriptsize
\rowcolors{2}{gray!25}{white}
\caption{Intelligent software web agents collaborative function perspectives.}
\begin{tabular} 
{p{4cm} p{3cm} p{3cm} p{3cm} p{3cm}}
\toprule
%\rowcolor{gray!50}
&
\textbf{Interoperability} &
\textbf{Communication} &
\textbf{Brokering  services} &
\textbf{Inter-agent coordination}   \\
\bottomrule  
\citet{bonatti2010reactive}  & -  & policies \& trust & - & - \\
\citet{berners2001semantic} & - & ontologies,  co-ordination \&  collaboration & - & - \\
\citet{bryson2002toward} & -  & protocols & - & internal co-ordination \\
\citet{ermolayev2004towards} & web services  & standard languages \& vocabularies & - & ontologies \\
\citet{garcia2008combining}    \citet{garcia2009ontology} & web services  & ontologies & data,  process \& function mediation & ontologies \\
\citet{gibbins2003agent}  & -  & standard languages \& vocabularies & system architecture & - \\
\citet{gladun2009application} & standards for interoperability & - & - & - \\
\citet{harth2017specifying} & linked data standards & - & - & - \\
\citet{hendler2001agents} & web services  & ontologies & logical descriptions & - \\
\citet{huhns2002agents}  & -  & protocols \&  standard languages \& vocabularies & enhanced directory services & autonomy vs coordination \\
\citet{kafer2018rule} &  linked data standards & - & - & - \\
\citet{motta2003irs} & web services  & protocols &  interaction framework & - \\
\citet{mcilraith2001semantic}  & -  & - & interaction framework & - \\
\citet{paolucci2003autonomous} & web services  & protocols & - & coordinating role \\
\citet{schraudner2020http} & linked data standards & indirect communication & - & - \\
\citet{shafiq2006bridging} & web services \&  agents & protocols & - & - \\
\citet{sycara2004dynamic} & web services \&  agents & ontologies & interaction framework & matchmaking \& brokering \\
\citet{tamma2008semantic} &  standards for interoperability &  ontological equivalence \&   reconciliation & - & - \\
\citet{tonti2003semantic} & - & communication policy & - & - \\

\bottomrule   
\end{tabular}
\normalsize
\label{tab:sq-collaborate}
\end{table*}

\subsection{Collaborate Functions} 

In the following, we further elaborate on various works that fall under the collaborative functions heading. \emph{Table}~\ref{tab:sq-collaborate} presents existing proposals for intelligent software web agents that are particularly relevant for both agent to human and agent to agent interactions, as well as internal interactions between agent components.

\paragraph{Interoperability}

Several authors \cite{hendler2001agents, ermolayev2004towards, paolucci2003autonomous, sycara2004dynamic, garcia2008combining, garcia2009ontology, motta2003irs} focus on interoperability from a web service perspective, putting a particular emphasis on automatic discovery, execution, selection, and composition.  However, only \citet{ garcia2008combining, garcia2009ontology} distinguish between data, process, and functionality interoperability. 
Both \citet{gladun2009application} and \citet{tamma2008semantic} highlight the need for standardisation when it comes to the interoperability in multi-agent systems. In particular, \citet{tamma2008semantic} differentiate between syntactic, semantic, and semiotic interoperability. 
While, \citet{shafiq2006bridging} focus specifically on communication between software agents and semantic web services by proposing an architecture that allows for interoperability via middleware that performs the necessary transformations.
More recently,  \citet{harth2017specifying, kafer2018rule} and \citet{schraudner2020http} have proposed agent architectures that are heavily reliant on linked data standards, which are interoperable by design.

\paragraph{Communication}

\citet{berners2001semantic} discusses the difficulties encountered when it comes to co-ordination and communication internationally.
\citet{huhns2002agents},  \citet{bryson2002toward}, \citet{paolucci2003autonomous},  \citet{shafiq2006bridging}, and \citet{motta2003irs} highlight the role played by various protocols (e.g., Web Services Description Language (WSDL),  Universal Description, Discovery
and Integration (UDDI), and Simple Object Access Protocol (SOAP)), when it comes to web service publishing, finding, and binding.
\citet{berners2001semantic, hendler2001agents, sycara2004dynamic}, and \citet{garcia2008combining, garcia2009ontology} propose the use of shared vocabularies in the form of ontologies for communication between service providers and consumers. \citet{garcia2008combining, garcia2009ontology} extend their use in order to cater for communication between architectural components.  Both of which raise issues from an interoperability perspective, especially in relation to ontological equivalence and reconciliation, as argued by \citet{tamma2008semantic}. 
When it comes to communication between agents,  \citet{ermolayev2004towards}, \citet{gibbins2003agent}, and \citet{huhns2002agents} highlight the need to standardise communication languages and vocabularies, in order to facilitate communication between agents. 
While, \citet{bonatti2010reactive} discuss the role played by policies and trust with a particular focus on negotiation.
From a communication management perspective, the communication architecture proposed by \citet{schraudner2020http} ensures that agents can only communicate with each other indirectly via the environment, whereas \citet{tonti2003semantic} use policies to control communication between agents.

\paragraph{Brokering services}

\citet{hendler2001agents} highlight that adding logical descriptions to web services will facilitate automated match making and brokering.
\citet{mcilraith2001semantic}, \citet{motta2003irs}, and \citet{sycara2004dynamic} propose frameworks whereby agent brokers are used to manage the interaction between service providers and consumers. 
While, \citet{gibbins2003agent} propose a system architecture and discuss its effectiveness via a proof of concept simulator based application.
\citet{huhns2002agents} in turn discusses how directory services could be enhanced via brokerage services that help to refine the number of potential sources that need to be consulted. 
\citet{garcia2008combining, garcia2009ontology} highlight the importance of interoperability when it comes to the brokering process, which is further subdivided into data, process, and functional mediation.

\paragraph{Inter-agent coordination}

\citet{huhns2002agents} highlights the tensions between autonomy and coordination, as it is necessary to relinquish some autonomy in order to honour commitments. 
The architecture proposed by \citet{paolucci2003autonomous} distinguishes between peers and super peers, the latter being responsible for coordinating several peers. 
While, \citet{bryson2002toward} argue that there is also the need to have co-ordination internally, for instance between software modules, such that it is possible to develop composite services. 
Both \citet{ermolayev2004towards} and \citet{garcia2008combining, garcia2009ontology} propose the use of common vocabularies in the form of ontologies for both inter-agent communication and co-ordination. 
While,  \citet{sycara2004dynamic} highlight the key roles played by matchmaking and brokering when it comes to multi-agent co-ordination. 

\begin{table*}[!t]
\scriptsize
\rowcolors{2}{gray!25}{white}
\caption{Intelligent software web agents code of conduct function perspectives.}
\begin{tabular} 
{p{4cm} p{2.3cm} p{2.3cm} p{2.3cm} p{2.3cm} p{2.3cm}}
\toprule
%\rowcolor{gray!50}
&
\textbf{Identity} &
\textbf{Security} &
\textbf{Privacy} &
\textbf{Trust} &
\textbf{Ethics} 
  \\
\bottomrule  
\citet{artz2007survey} & credentials  \& policies & credentials  \&  policies &  policies &  trust  judgements & - \\
\citet{bao2007privacy} & - & - & privacy  preserving reasoning & - & -\\
\citet{berners2001semantic} & URIs & -  & - & digital signatures  & - \\
\citet{casanovas2015semantic} & - & -  & - & - & regulatory models \\
\citet{chen2004intelligent} & - & ontologies \& policies & ontologies \& policies  & trust \& privacy & - \\
\citet{ermolayev2004towards} & URIs & - & - & credibility  \& trust  & -\\
\citet{garcia2008combining}  \citet{garcia2009ontology} & URIs  & - & - & - & -\\
\citet{gandon2004semantic} & credentials  \& policies  & access policies  & privacy architecture & - & -\\
\citet{harth2017specifying} & URIs  & - & - & - & -\\
\citet{hendler2001agents} & URIs & -  & - & proofs & - \\
\citet{jutla2004privacy} & - & - & privacy ontology & trust \& privacy & -\\
\citet{jutla2006pecan} & -  & - & privacy architecture &  trust \& privacy & - \\
\citet{kafer2018rule} & URIs  & - & - & - & -\\
\citet{kagal2004authorization} & - & digital signatures \& encryption & - & - & -\\
\citet{kirraneintelligent} & credentials  \& policies & access policies  & - & trust \& provenance & usage policies \\
\citet{kravari2016policy} & - & - & privacy contracts & trust via contract & - \\
\citet{oren2008sindice} & URIs & - & - & - & social conventions e.g., robots.txt \\
\citet{palmirani2018pronto} & - & - & privacy  ontology & - & -\\
\citet{sycara2004dynamic} & - & matchmakers \& security & matchmakers \& privacy & - & - \\ 
\citet{tamma2008semantic} & URIs & - & - & - & -\\
\citet{tonti2003semantic} & - & - & - & trust \& policies \\
\bottomrule   
\end{tabular}
\normalsize
\label{tab:sq-governance}
\end{table*}

\subsection{Code of Conduct Functions}

The functions summarised in \emph{Table}~\ref{tab:sq-governance} and further elaborated on below are particularly relevant for the controller component, however they may also impact the design of several other components, thus they need to be considered when it comes to the architectural design of the system.

\paragraph{Identification}

Several authors \cite{berners2001semantic, hendler2001agents, ermolayev2004towards, oren2008sindice, garcia2008combining, garcia2009ontology, tamma2008semantic, harth2017specifying, kafer2018rule} highlight the role played by Uniform Resource Identifiers (URIs) when it comes to the identification of resources (e.g., web services, ontologies, agents).  While, \citet{artz2007survey}, \citet{gandon2004semantic} and \citet{kirraneintelligent} focus on the authentication of actors using credentials in the form of digital signatures together with policies.  

\paragraph{Security}

Both \citet{gandon2004semantic} and \citet{kirraneintelligent} identify the need for access control policy specification and enforcement.  Although, \citet{chen2004intelligent} argue that together ontologies and declarative policies can be used for both privacy and security,  they do not go into specific details on their use from a security perspective.  \citet{artz2007survey} discuss security in terms of using credentials and policy languages used in order to determine trust in an entity.  While, \citet{kagal2004authorization} propose ontologies that can be used to sign and encrypt messages exchanged between service providers and consumers. \citet{sycara2004dynamic} in turn argue that matchmakers can be used to address security concerns by offering a choice of providers.

\paragraph{Privacy}

\citet{chen2004intelligent} discuss how their context broker architecture can be used to control the sharing and use of personal data.  \citet{jutla2006pecan} propose an agent based architecture that can be used to allow the specification and enforcement of privacy preferences.  \citet{gandon2004semantic} in turn propose an agent based architecture that protects and mediates access to personal resources.  Both \citet{jutla2004privacy} and \citet{palmirani2018pronto} propose high level ontologies that can be used to specify privacy protection mechanisms in the form of laws, standards, societal norms, and guidelines. In addition, the authors describe how the proposed ontologies could be used by privacy agents to identify privacy issues. Whereas, \citet{bao2007privacy} propose a framework that can be used for privacy preserving reasoning when only partial access to data is permitted.  \citet{artz2007survey} discuss privacy from a trust negotiation perspective and point to several policy languages that can be used to protect privacy.  \citet{kravari2016policy} also focus on enhancing privacy via trust,  proposing a policy-based e-Contract workflow management methodology. 
\citet{sycara2004dynamic} identify matchmakers as a means to cater for better privacy by offering a choice of providers.

\paragraph{Trust}

\citet{berners2001semantic} discuss the role played by digital signatures when it comes to verifying that information has been provided by trusted sources. While, \citet{hendler2001agents} focuses more broadly on using proof exchange to facilitate trust.  Additionally, the detailed survey on trust models and mechanisms at the intersection of trust and the semantic web,  conducted by \citet{artz2007survey},  is motivated by the need for agents to make trust judgements based on available data that may vary in terms of quality and truth.  The web service composition framework proposed by \citet{ermolayev2004towards} considers both the credibility and trustworthiness of service providers as a key requirement that needs to be considered.  While,  \citet{chen2004intelligent}, \citet{jutla2004privacy}, and \citet{jutla2006pecan} examine trust from a personal data processing perspective, proposing a system that can be used to determine if the users privacy preferences are adhered to.  \citet{kirraneintelligent} highlight the link between trust, transparency and provenance and point to several potential starting points.  
\citet{kravari2016policy} propose a policy-based e-Contract workflow management methodology that can be used to establish trust between agents and service providers. While, \citet{tonti2003semantic} highlight the role between trust and policies from a trust management perspective.

\paragraph{Ethics}

When it comes to intelligent software agents, \citet{casanovas2015semantic} argues that there is a need for both normative and institutional regulatory models in order to not only reason over legal norms, but also judicial and political decision making, best practices, ethical principles and values. \citet{oren2008sindice} focus specifically on good behaviour when it comes to crawling data, stating it is important to respect the robot.txt access restrictions and to be mindful of the resource limitations of the data provider. \citet{kirraneintelligent} identify usage restrictions in the form of access policies, usage constraints, regulatory constrains, and social norms as key requirements needed to realise the intelligent web agent vision.

\begin{table*}[!t]
\scriptsize
\rowcolors{2}{gray!25}{white}
\caption{Intelligent software web agents robustness function perspectives.}
\begin{tabular} 
{p{5cm} p{6cm} p{6cm}}
\toprule
%\rowcolor{gray!50}
&
\textbf{Stability,  Performance \&  Scalability}  &
\textbf{Verification}  
 \\
\bottomrule  
\citet{gandon2004semantic}  & - &  empirical evaluation\\
\citet{garcia2008combining}    \citet{garcia2009ontology}  & evaluation plans & -\\
\citet{ghanadbashi2021using} & - & simulation \\
\citet{jochum2019data} & -  & ruleset correctness \\
\citet{jutla2006pecan} & borrowing/extending  metrics &  -\\
\citet{kafer2018rule} & -  &  simulation \\
\citet{ksystra2016formal} & - & safety properties \\
\citet{leite2013using,leite2014case} & -  & comparative analysis \\
\citet{merkle2019cooperative} & - & learning tasks \\
\citet{schraudner2020http} & - &  simulation \\
\citet{scioscia2014mobile} &  reasoning engine  performance \&  memory usage &  reasoning correctness \\
\citet{shafiq2006bridging} & service lookup \& invocation performance & -\\
\citet{sycara2004dynamic} & several different performance characteristics & - \\
\citet{van2015semantic} & - & norm compliance checking \\
\bottomrule   
\end{tabular}
\normalsize
\label{tab:sq-performance}
\end{table*}

\subsection{Robustness Functions}

Finally, the existing work at the intersection of intelligent software web agents and robustness, summarised in \emph{Table}~\ref{tab:sq-performance}, is useful for both assessing the maturity of the exiting proposals, and comparing and contrasting different technological choices.

\paragraph{Stability,  Performance \& Scalability} 

\citet{shafiq2006bridging} examine the performance of both their service lookup via UDDI and service invocation via WSDL, and conclude that lookups scale linearly with increasing parameters, while service invocation depends on the complexity of the input and output parameters.  Although \citet{jutla2006pecan} do not conduct a performance assessment they identify the need for borrowing/extending metrics from other domains, such as response time, throughput,  effectiveness, ease of use, and usefulness.
Likewise, \citet{garcia2008combining, garcia2009ontology} discuss the importance of performance assessment and provide detailed plans that they aim to execute in future work. 
While, \citet{scioscia2014mobile} used a reference dataset to compare their reasoning engine performance, in terms of both classification and satisfiability, to other well known reasoners, and the memory usage on a mobile device to that of a personal computer.  
\citet{sycara2004dynamic} take a broad view on performance identifying the need to assess different performance characteristics, such as privacy, robustness, adaptability,  and load balancing.

\paragraph{Verification} 

\citet{scioscia2014mobile} used a reference dataset to assess the effectiveness of their reasoner on a classification task, in terms of correctness, parsing errors, memory exceptions,  and timeouts.  While, \citet{gandon2004semantic} perform an empirical evaluation of their architecture via a campus community agent application, where participants were asked to perform various tasks with a view to obtaining feedback on the effectiveness of the system. 
\citet{leite2013using} and \citet{leite2014case} demonstrate the effectiveness of their proposals by performing a comparitive analysis to other hybrid agent architectures.  Additionally, a number of authors evaluated their proposals using smart home \citep{kafer2018rule},  production environment \citep{schraudner2020http}, and real time traffic  \citep{ghanadbashi2021using} simulations.
Other evaluations included ruleset correctness \citep{jochum2019data}, safety \citep{ksystra2016formal},  and norm compliance checking \citep{van2015semantic}.

%% file: content/iswaarchitecture.tex
% !TEX root = ../main.tex
\section{Intelligent Software Web Agents: Architectural Components}\label{sec:architecture}

%The goal of this section is to propose a hybrid semantic web agent architecture that can be used to guide future work on intelligent software web agents. We start by describing the high level architecture and discussing the role played by existing semantic web standards.  Following on from this, we revisit the use case requirements and status quo analysis in order to map requirements and perspectives to the architectural components, and to highlight existing research challenges and opportunities.  
%\subsection{A Hybrid Semantic Web Agent Architecture}

\begin{figure*}[t!]
\center
  \includegraphics[width=0.8\textwidth]{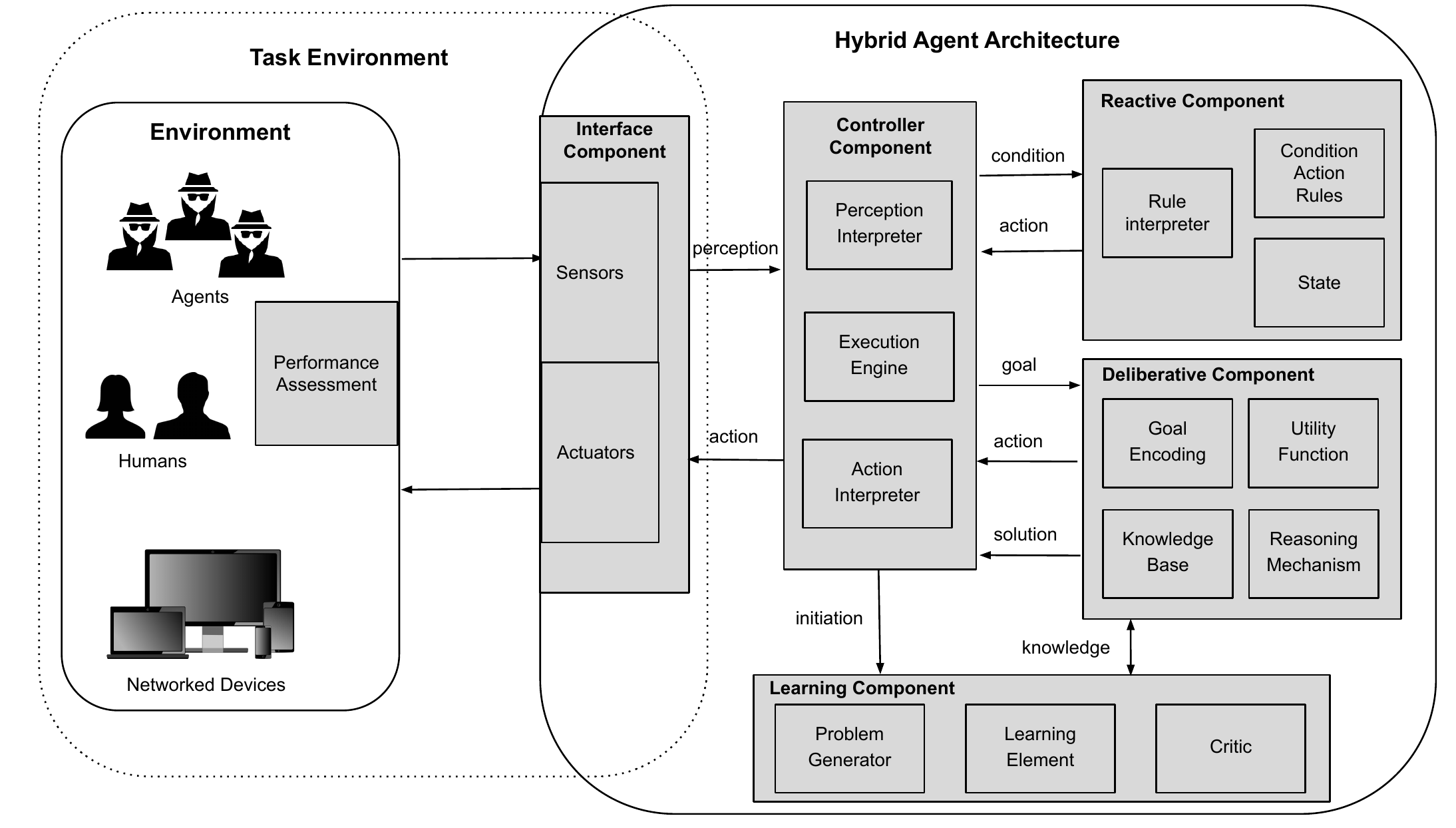}
  \caption{A hybrid agent architecture.}
  \label{fig:architecture}
\end{figure*}

In the following, we propose a hybrid semantic web agent architecture,  and discuss how the architecture components introduced in \emph{Section}~\ref{sec:background}, together with semantic web standards and community activities, could potentially be used to realise our information, booking, and planning agents. 
Rather than proposing three different architectures we propose a single architecture with optional components. 
The proposed hybrid agent architecture, which is depicted in \emph{Figure}~\ref{fig:architecture}, provides support for realtime interaction via its reactive component and sophisticated reasoning via its deliberative component, both of which are necessary in order to realise our scheduling agent.  

\begin{table*}[!t]
\scriptsize
\rowcolors{2}{gray!25}{white}
\caption{Intelligent software web agents interface component.}
{\color{black}\begin{tabular} 
{p{3.2cm} p{4.9cm} p{8.9cm}}
\toprule
&
\textbf{Standards} &
\textbf{Community Activities} 
\\ 
\bottomrule  
Languages,  Ontologies \& \newline Vocabularies  & HTML, CSS,  WSML, WSDL, FIPA ACL, FIPA RDF CL, OWL, OWL-S,  SAWSDL, WSMO & FIPA RDF CL extension \citep{pai2016semantic},  SEA\_ML \citep{challenger2018development},  compare OWL-S, WSMO and SAWSDL \citep{wang2015survey}, FIPA process ontology \citep{gibbins2003agent},  OWL-S \& policies \\
%%%
Models \& Frameworks   & RDF, UDDI, REST & SWS discovery \& composition \citep{seghir2015decentralized, venkatachalam2016comprehensive, seghir2017semantic, seghir2017new},   SWS \& JADE \citep{zapater2015semantic}, SWS design methodology \citep{hajji2017semantic}, hypermedia controls \citep{verborgh2015bottom}  \\
%%%
Protocols  & HTTP,  SOAP, LDN  &  WS publication protocol \citep{seghir2017semantic}, LDN agent communication protocol \citep{calbimonte2017multi}, ACL/SOAP converter \citep{shafiq2006bridging}\\
\bottomrule   
\end{tabular}}
\normalsize
\label{tab:interface-comp}
\end{table*}

\subsection{Interface Component}
\label{sec:isa-interface}

In our use case scenario, \texttt{Sensors} and \texttt{Actuators} take the form of either web interfaces, rendered via networked devices used for agent to human interaction, or web services, residing on networked devices used for agent to agent interactions.  
\emph{Table}~\ref{tab:interface-comp} summarises the relevant W3C standardisation efforts and community activities discussed in detail below.

\bigskip

\paragraph{Sensors \& Actuators}

The Hypertext Transfer Protocol (HTTP)\footnote{\url{https://tools.ietf.org/html/rfc7231}} is an application level protocol that forms the basis for data communication via the web.
Communication involves a simple request/response protocol that can be used for data exchange. 
The Linked Data Notifications (LDN)\footnote{\url{https://www.w3.org/TR/ldn/}} specification in turn describes how HTTP together with the Resource Description Framework (RDF)\footnote{\url{https://www.w3.org/TR/rdf11-concepts/}} can be used by senders to push messages to recipients. 
When it comes to serving web content there are numerous web servers to choose from (c.f., NGINX\footnote{\url{https://www.nginx.com/}},  Apache Tomcat\footnote{\url{http://tomcat.apache.org/}}). 
From a web interface perspective,  the Hypertext Markup Language (HTML)\footnote{\url{https://www.w3.org/TR/html52/}} and Cascading Style Sheets (CSS)\footnote{\url{https://www.w3.org/TR/CSS2/}} can be used to develop responsive web applications that enable humans to interact with intelligent software agents.
From a web services perspective, the Simple Object Access Protocol (SOAP)\footnote{\url{https://www.w3.org/TR/soap12-part1/}} and the Representational State Transfer (REST)\footnote{\url{https://www.ics.uci.edu/~fielding/pubs/dissertation/rest\_arch\_style.htm}}
architecture style are the predominant Application Programming Interface (API) approaches used in practice.
%, both of which can run over the Hypertext Transfer Protocol (HTTP)\footnote{\url{https://tools.ietf.org/html/rfc7231}}.  
%
Web service discovery is supported via registries and indexes, whereby protocols such as the Universal Description, Discovery and Integration (UDDI)\footnote{\url{http://uddi.xml.org/specification}} can be used to publish and discover web services \cite{booth2004web}.
There are also several standardisation initiatives relating to semantic web services that use formal ontology-based annotations to describe the service in a manner that can be automatically interpreted by machines (c.f., the Web Ontology Language for Web Services (OWL-S)\footnote{\url{https://www.w3.org/Submission/OWL-S/}}, the Web Service Modeling Language (WSML)\footnote{\url{https://www.w3.org/Submission/WSML/}}, the W3C standard Semantic Annotations for Web Services Description Language (WSDL) and XML Schema (SAWSDL)\footnote{\url{https://www.w3.org/TR/sawsdl/}}).  
When it comes to agent specific standardisation efforts, the Foundation for Intelligent Physical Agents (FIPA) propose several standards that support agent to agent communication \citep{poslad2007specifying}, such as the FIPA Agent Communication Language (ACL)\footnote{\url{http://www.fipa.org/specs/fipa00061/index.html}} and the FIPA RDF Content Language Specification\footnote{\url{http://www.fipa.org/specs/fipa00011/XC00011B.html}} which describes how RDF can be used to encode the message content.

%%%%%%%%%%%%
% Research
%%%%%%%%%%%%

Additionally,  there have been numerous works that focus on using enhancing,  and supplementing existing standards, from an intelligent software web agent perspective.  
In terms of agent specific languages,  \citet{wang2015survey} compare OWL-S,  Web Service Modeling Ontology (WSMO)\footnote{\url{https://www.w3.org/Submission/WSMO/}} and SAWSDL from the providers, requesters, and brokers perspectives.  On the other hand \citet{pai2016semantic} propose a lightweight ontology-based content language based on the FIPA RDF Content Language (CL).  While, \citet{challenger2018development} introduce a semantic web enabled agent modelling Language (SEA\_ML), which they apply in the context of an  E-barter system.  \citet{gibbins2003agent} propose a process ontology, inspired by the FIPA agent communication language, that describes various messages types that can be used to describe web services.  Whereas, \citet{kagal2004authorization} demonstrate how policies can be embedded into OWL-S descriptions.

%%%
When it comes to models and frameworks, \citet{venkatachalam2016comprehensive} provide a comprehensive survey of existing work on semantic web service (SWS) composition and discovery.  More recent works primarily focus on using ontologies to semantically described RESTful web services \citep{dantas2015semantic}, new approaches for service discovery that leverage user profiles and metadata catalogs \citep{seghir2015decentralized, seghir2017semantic, seghir2017new}, and proposing methodologies that support the modelling and design of SWSs \citep{hajji2017semantic}.  From an implementation perspective, \citet{zapater2015semantic} demonstrate how the JADE Multi-agent System (MAS) development platform can be enhanced with service discovery capabilities. \citet{verborgh2015bottom} in turn argue that there is a need to construct Web APIs out of reusable building blocks and for the use of hypermedia controls to describe both the functional and non-functional aspects of the service.  

%%%
From a protocol perspective,  \citet{seghir2017semantic} propose web service publication and discovery protocols that are represented in the form of sequence diagrams. 
While, \citet{shafiq2006bridging} propose an abstract architecture, combining web service and FIPA standardisation efforts, which is capable of translating FIPA ACL to SOAP and visa versa.
Others have demonstrated how LDNs can be extended to cater for agent communication \citep{calbimonte2017multi}.

%%%%%%%%%%%%
%END NEW
%%%%%%%%%%%%

\begin{table*}[!t]
\scriptsize
\caption{Intelligent software web agents reactive component.}
{\color{black}\begin{tabular} 
{p{3.2cm} p{4.9cm} p{8.9cm}}
\toprule
&
\textbf{Standards} &
\textbf{Community Activities} 
\\ 
\bottomrule  
\rowcolor{gray!50}
\multicolumn{3}{l}{\textbf{Condition Action Rules}}  \\
\rowcolor{white}
Languages,  Ontologies \& \newline Vocabularies  & PRR,  RuleML, RIF, SWRL, RDF, XML &  condition action rules \citep{harth2017specifying, kafer2018rule},  event condition action rules \citep{boley2001design, ksystra2016formal, papamarkos2003event, papamarkos2003event, papamarkos2004rdftl, poulovassilis2006event, jochum2019data}, event condition transaction language \citep{gomes2015procedure}\\
\rowcolor{gray!25}
Models \&  \newline Frameworks   &  RDF &  standards based system architectures \citep{harth2017specifying,schraudner2020http}, formal verification framework \citep{ksystra2016formal}  \\
\hline
\rowcolor{gray!50}
\multicolumn{3}{l}{\textbf{State}}  \\
\rowcolor{white}
Languages,  Ontologies \& \newline Vocabularies  &  RDFS, OWL LD &  RDF, RDFS, and OWL LD \citep{harth2017specifying},  RDF graphs \citep{papamarkos2003event, papamarkos2004rdftl, poulovassilis2006event}\\
%%%
\rowcolor{gray!25}
Models \& Frameworks   &  RDF &  memory, commitments, claims, goals,  and intentions \citep{boley2001design},  sequence of states \citep{gomes2015procedure},  active, inactive \& done workflow state \citep{jochum2019data},  check violated state using model checking \citep{ksystra2016formal} \\
%%%
\hline
\rowcolor{gray!50}
\multicolumn{3}{l}{\textbf{Rule Interpreter}}  \\
\rowcolor{white}
Languages,  Ontologies \& \newline Vocabularies  &  RuleML engine, SPARQL & RuleML design rationale  \citep{boley2001design},  SPARQL enabled interpreter   \citep{harth2017specifying},  parser and translator \citep{papamarkos2003event, papamarkos2004rdftl, poulovassilis2006event} \\
%%%
\rowcolor{gray!25}
Models \& Frameworks   & RDF &  workflow meta model \citep{jochum2019data},  Protune policy engine \citep{bonatti2010reactive}\\
%%%
\bottomrule   
\end{tabular}}
\normalsize
\label{tab:reactive-comp}
\end{table*}

\subsection{Reactive Component}
\label{sec:isa-reactive}

The \texttt{Reactive Component} takes as input a condition and returns an action based on a set of \texttt{Condition Action Rules}. More sophisticated reactive components use \texttt{State} to further refine the conditions used to determine the action that is required.   
\emph{Table}~\ref{tab:reactive-comp} summarises existing work both in terms of standardisation and community activities.

\paragraph{Condition Action Rules} 

When it comes to the specification of condition action rules there are several applicable standardisation efforts.
The Production Rule Representation (PRR)\footnote{\url{https://www.omg.org/spec/PRR/About-PRR/}} specification, developed by the Object Management Group,  provides a standard mechanism for encoding rules of the form IF \emph{condition} THEN \emph{action} statements. 
The Rule Markup Language (RuleML)\footnote{\url{http://wiki.ruleml.org/index.php/Specification_of_RuleML}} is a family of languages that provide support for the specification and interchange of both derivation and reaction rules \cite{boley2010ruleml}.  
The W3C  Rule Interchange Format (RIF)\footnote{\url{https://www.w3.org/TR/rif-overview/}} in turn is an interchange format that can be used to exchange rules between different rule systems.
The RIF Production Rule Dialect\footnote{\url{https://www.w3.org/TR/rif-prd/}} caters specifically for production rules.
While,  the W3C Semantic Web Rule Language (SWRL)\footnote{\url{https://www.w3.org/Submission/SWRL/}} is a language that combines rules and logic,  for a subset of RuleML and a subset of the Web Ontology Language (OWL)\footnote{\url{https://www.w3.org/TR/2012/REC-owl2-overview-20121211/}}.

Over the years, researchers have proposed a variety of reactive rules languages that allow for the specification of condition action rules \citep{harth2017specifying, kafer2018rule}, event condition action rules \citep{boley2001design, ksystra2016formal, papamarkos2003event, papamarkos2004rdftl, poulovassilis2006event, jochum2019data}, and event condition transaction rules that combines condition action rules with transaction logic \citep{gomes2015procedure}.  When it comes to models and frameworks, \citet{harth2017specifying} and \citet{schraudner2020http} propose W3C standard based architectures.  While, \citet{ksystra2016formal} propose a formal framework for analysing reactive rules in order to safeguard against unpredictable behaviour.

\paragraph{State}
  
A reactive agent with state maintains knowledge about the world and the current state of the environment.  RDF is a general purpose language that could be used to represent information in a machine interpretable format.  
The PRR specification describes a metamodel for encoding production rules using the Extensible Markup Language (XML) Metadata Interchange\footnote{\url{https://www.omg.org/spec/XMI/2.5.1/PDF}},  which is abstract in nature.
XML is the native encoding for RuleML, however a JavaScript Object Notation (JSON) serialisation is also provided.
Although RIF supports several different encodings, XML is the primary medium of exchange between different rule systems. 
The SWRL specification uses an abstract Extended Backus-Naur Form (EBNF) syntax, which can easily be encoded in XML and/or RDF.

The agents envisaged by \citet{harth2017specifying} model state using RDF, RDFS, and OWL LD.  While, \citet{papamarkos2003event,papamarkos2004rdftl, poulovassilis2006event} use state to refer to RDF graphs. 
\citet{boley2001design} enumerate several categories that encompass the mental state of an agent,  from a memory, commitments, claims, goals,  and intentions perspective.  
Considering their transactional focus, \citet{gomes2015procedure} work with a sequence of knowledge base states, which they refer to as paths.  
The system proposed by \citet{jochum2019data} proposes three workflow states: active, inactive and done. 
In their formal verification framework \citet{ksystra2016formal} use model checking to find violated states.
Whereas, the basic agent architecture proposed by \citet{schraudner2020http} does not maintain state. 

\paragraph{Rule Interpreter}
 
Although some rule engines provide support for RuleML and SWRL rules (c.f., RDFox\footnote{\label{RDFoX}\url{https://www.oxfordsemantic.tech/}}),  when it comes to condition action rules, a simple interpreter that is able to match conditions would suffice. The rule interpreter is responsible for finding rules whose conditions are satisfied, and for triggering the corresponding actions.  In the case of conflicting rules, a conflict resolution mechanism is required.  

\citet{boley2001design} elaborate on the design rationale underpinning RuleML, which provides support for reactive rules, derivative rules, and integrity constraints.
\citet{poulovassilis2006event} propose an event condition action rule language that can be applied to RDF data,  entitled RDF Triggering Language (RDFTL) in the form of an RDF repository wrapper that leverages the repositories querying capabilities. 
The interpreters used by \citet{papamarkos2003event, papamarkos2004rdftl} and \citet{poulovassilis2006event} includes a parser that performs syntactic validation and a translator that translates queries such that they can be executed by the underlying RDF store. 
In the system proposed by \citet{harth2017specifying} conditions are checked against state, by a SPARQL enabled interpreter,  with optional support for some simple RDFS and OWL LD based reasoning. 
\citet{bonatti2010reactive} in turn propose a reactive policy framework, called Protune, that can be used to guide system behaviour.

\begin{table*}[!t]
\scriptsize
\caption{Intelligent software web agents deliberative component.}
{\color{black}\begin{tabular} 
{p{3.2cm} p{4.9cm} p{8.9cm}}
\toprule
&
\textbf{Standards} &
\textbf{Community Activities} 
\\ 
\bottomrule  
\rowcolor{gray!50}
\multicolumn{3}{l}{\textbf{Knowledge Base}} \\
\rowcolor{white}
Languages,  Ontologies \& \newline Vocabularies  & RDFS, OWL,  ODRL  &  belief-augmented OWL \citep{dong2015bowl}, normative language ontology \citep{fornara2018using,fornara2019using} \\
%%%
\rowcolor{gray!25}
Models \& Frameworks   &  RDF,  ODRL & ODRL policy activation \& temporal validity \citep{fornara2018using,fornara2019using}, belief desire intention principles  \citep{challenger2018development},  Jason MAS Platform  \&  ontological knowledge \citep{demarchi2018integration},  ontology integration \& automatic reconciliation \\
%%%
\hline
\rowcolor{gray!50}
\multicolumn{3}{l}{\textbf{Reasoning Engine}}  \\
\rowcolor{white}
Languages,  Ontologies \& \newline Vocabularies  & RDFS, OWL,  ODRL,  SPARQL 1.1 Entailment Regimes  & OWL reasoning \citep{acar2015udecide},  rule based reasoning\citep{ming2017ontology}, reasoning over obligations \& permissions \citep{fornara2018using,fornara2019using}\\
%%%
\rowcolor{gray!25}
Models \& Frameworks   & RDF, ODRL   &  belief desire intention reasoning \citep{challenger2018development, demarchi2018integration},  reasoning over  incomplete, subjective \& inconsistent data \citep{dong2015bowl}  \\
\hline
\rowcolor{gray!50}
\multicolumn{3}{l}{\textbf{Goal Encoding}}  \\
\rowcolor{white}
Languages,  Ontologies \& \newline Vocabularies  &  RDFS, OWL &  roles, behaviors,
plans, beliefs, and goal concepts \citep{challenger2018development}, task, planing \& scheduling ontologies \citep{mizoguchi1995ontology, rajpathak2004ontological, pham2011practical},  \\
%%%
\rowcolor{gray!25}
Models \& Frameworks    & RDF  &  constraint logic programming goals \citep{dong2015bowl}  \\
\hline
\rowcolor{gray!50}
\multicolumn{3}{l}{\textbf{Utility Function}}  \\
\rowcolor{white}
Languages,  Ontologies \& \newline Vocabularies  & RDFS, OWL &  degree of inclination \citep{dong2015bowl},  utility values assigned to classes using the uDecide protégé plugin  \citep{acar2015udecide} \\
%%%
\rowcolor{gray!25}
Models \& Frameworks & RDF  &  utility theory based modelling \citep{ming2017ontology},  e-bartering economics  \citep{challenger2018development}  \\
\bottomrule   
\end{tabular}}
\normalsize
\label{tab:deliberative-comp}
\end{table*} 
 
\subsection{Deliberative Component}
\label{sec:isa-deliberative}

The \texttt{Deliberative Component} takes as input a goal and either returns a solution or an action (that needs to be carried out before a solution can be determined).  The \texttt{Knowledge Base} is used to store the knowledge the agent has about the world. The \texttt{Goal Encoding} is responsible for intercepting the request and updating the knowledge base accordingly. Together the \texttt{Reasoning Engine} and the \texttt{Utility Function} are responsible for deriving a solution or further actions that need to be fed back to the execution engine.  
\emph{Table}~\ref{tab:deliberative-comp} summarises relevant standardisation efforts that could be used to realise this component and various community activities that make use of them.

\bigskip

\paragraph{Knowledge Base} 
 
A deliberative agent maintains knowledge about the world and the current state of the environment in its knowledge base.  In the case of our intelligent software web agents, RDF is used to represent information about resources accessible via the web.  The RDF Schema\footnote{\url{https://www.w3.org/TR/rdf-schema/}} specification defines a set of classes and properties used to describe RDF data.  However, using RDFS, it is not possible to represent complex statements that include cardinality constraints, or to model complex relations between classes, such as disjointness or equivalence.  The OWL Web Ontology Language\footnote{\url{https://www.w3.org/TR/owl2-overview/}} standard caters for the encoding of relations between classes, roles and individuals,  while at the same time providing support for logical operations and cardinality constraints.  Additionally, the Open Digital Rights Language (ODRL)\footnote{\url{https://www.w3.org/TR/odrl-model/}} could potentially be used to represent other constraints in the form of policies and norms.

\citet{dong2015bowl} combine OWL with belief augmented frames based logic, which can be used to model evidence for or against a statement. 
While, \citet{fornara2018using} and \citet{fornara2019using} describe an OWL based normative language ontology and demonstrate how their ODRL extension caters for policy activation and temporal relevancy can be used by agents to reason about obligations and permissions.
The agents envisaged by \citet{challenger2018development} are modelled based on the belief-desire intention (BDI) principles \citep{rao1995bdi}.
Whereas, \citet{demarchi2018integration} demonstrates how the Jason multi-agent system development platform \citep{bordini2005bdi} can be amended to make use of ontological knowledge available on the web in order to update the agents knowledge base.  
From an interoperability perspective, \citet{lister2002reconciling} propose several alternative strategies that could be used for automatic ontology reconciliation.

\paragraph{Reasoning Engine} 

OWL2\footnote{\url{https://www.w3.org/TR/owl2-overview/}} comes in two flavours: OWL2 Full and OWL2 DL.  In turn, OWL2 DL is composed of three profiles (OWL2EL, OWL2QL and OWL2RL) that are based on well used DL constructs.  The syntactic restrictions imposed on each profile are used to significantly simplify ontological reasoning. 
OWL 2 EL is designed for applications that require very large ontologies., whereby polynomial time reasoning is achieved at the cost of expressiveness.
OWL 2 QL is particularly suitable for applications with lightweight ontologies and a large number of individuals, which need to be accessed via relational queries.
Finally,  OWL 2 RL provides support for applications with lightweight ontologies, and a large number of individuals that make use of rule based inference and constraints mechanisms.
Over the years the community has developed several reasoners that are capable of reasoning over OWL ontologies, albeit often with some restrictions (c.f.,  Pellet\footnote{\url{https://www.w3.org/2001/sw/wiki/Pellet}}, HermiT\footnote{\url{http://www.hermit-reasoner.com/}}, FACT++\footnote{\url{http://owl.cs.manchester.ac.uk/tools/fact/}}, Racer\footnote{\url{http://www.ifis.uni-luebeck.de/~moeller/racer/}},  and RDFox\footnoteref{RDFoX}).
Additionally, SPARQL 1.1 Entailment Regimes\footnote{\url{https://www.w3.org/TR/sparql11-entailment/}} can be used to consider implicit (i.e. inferred) data during query execution based on RDF entailment regimes.

Beyond simple OWL based \citep{acar2015udecide, dong2015bowl, fornara2018using,fornara2019using} and rule based \citep{ming2017ontology,pham2011practical} reasoning,  researchers have demonstrated the potential for reasoning over beliefs, desires and intentions \citep{challenger2018development,demarchi2018integration},  policies and norms \citep{fornara2018using,fornara2019using}, and incomplete, subjective,  and inconsistent data \citep{dong2015bowl}.

\paragraph{Goal Encoding} 

Given a goal, or set of goals, the agent uses the current state of the world together with the desired state of the world, deduced from its goal(s), in order to infer a solution or further actions that need to be performed.  Here RDF, RDFS, and OWL can be used to encode the agents goal(s).  

Although there are no standard mechanisms for goal encoding, 
the domain-specific modelling language proposed by \citet{challenger2018development} covers goals and other predominant agent concepts (i.e., roles, behaviors, plans,  and beliefs). Additionally, several researchers have proposed task, planing, and scheduling ontologies \citep{mizoguchi1995ontology, rajpathak2004ontological, pham2011practical}.  While, the belief framework proposed by \citet{dong2015bowl} uses constraint logic programming goals.

\paragraph{Utility Function}
  
The utility function is responsible for assessing possible solutions based on the desired state of the world, and the preferences defined by the person or agent that specifies the goal(s) and associated constraints. According to utility theory~\cite{morgenstern1953theory} informed decisions should be made by examining the goal(s), the actions needed to achieve the goal(s), and the various preferences from a greatest expected satisfaction perspective.  Here, a utility theory based modelling, such as that adopted by \citet{brown1998utility} and \citet{ming2017ontology},  could be used to guide the development of the utility function.  

\citet{dong2015bowl} define a utility function based on the difference between belief and disbelief values (i.e. the degree of inclination). While,  \citet{acar2015udecide} propose a protégé plugin called uDecide that can be used to assign utility values to classes that are subsequently used by the utility function in order to determine the optimal course of action.
The template-based ontological method proposed by \citet{ming2017ontology} is rooted in utility theory based modelling \citep{brown1998utility}. 
From a domain specific perspective, the semantic web enabled BDI multi-agent system proposed by \citet{challenger2018development} builds upon the utility function research specifically focused on the proposed e-bartering system.

\begin{table*}[!t]
\scriptsize
\caption{Intelligent software web agents learning component.}
{\color{black}\begin{tabular} 
{p{3.2cm} p{4.9cm} p{8.9cm}}
\toprule
&
\textbf{Standards} &
\textbf{Community Activities} 
\\
\bottomrule  
\rowcolor{gray!50} 
\multicolumn{3}{l}{\textbf{Problem Generator \& Critic}} \\
\rowcolor{white}
Languages,  Ontologies \& \newline Vocabularies  &  RDFS, OWL & OWL \& PDDL \citep{buoncompagni2017ros, kootbally2017overview} \\
\rowcolor{gray!25} 
Models \& Frameworks  & RDF & critic, learning element \& problem generator \citep{leite2013using},  adaptive behaviour \& norms \citep{van2015semantic, buoncompagni2017ros}  \\
\hline
\rowcolor{gray!50} 
\multicolumn{3}{l}{\textbf{Learning Element}} \\
\rowcolor{white}
Languages,  Ontologies \& \newline Vocabularies  & OWL & ontology learning \citep{wong2009learning,puerto2012automatic}, reinforcement learning \& policies/strategies \citep{merkle2019cooperative},  computationally complex \citep{albrecht2018autonomous} \\
\rowcolor{gray!25} 
Models \& Frameworks  \newline Models  & RDF &  critic, learning element \& problem generator \citep{leite2013using,leite2014case}, spatial information \& semantic web mining \citep{young2016towards}, automatic goal generation model \citep{ghanadbashi2021using},   linguistic,  statistic and logic based \citep{asim2018survey}\\
\bottomrule   
\end{tabular}}
\normalsize
\label{tab:learning-comp}
\end{table*} 

\subsection{Learning Component}
\label{sec:isa-learning}

The learning component, which is composed of the \texttt{Problem Generator}, \texttt{Learning Element}, and \texttt{Critic}, could be used to develop more advanced intelligent software web agents that are capable of learning from past experiences and thus becoming more effective over time.  This component interacts with both the \texttt{Controller Component} and the \texttt{Deliberative Component}. The former is responsible for initiating the learning process, while the latter is used to ascertain existing knowledge, perform learning based reasoning tasks, and store the outputs of the learning process. 
Considering that simple agents (such as the information and booking agents presented in \emph{Section}~\ref{sec:swa}) do not necessarily need learning capabilities, in the proposed architecture, following a typical separation of duties engineering practice, we separate the learning component from the deliberative component. That being said,  it is worth noting that there is a high level of interaction between these components.  Although the tools and techniques that could be used to support agent learning have not yet been considered from a standardisation perspective,  in \emph{Table}~\ref{tab:learning-comp} we summarise preliminary research that could form a starting point for potential standardisation discussions.  

\bigskip

\paragraph{Learning Element}

Although the W3C doesn't have any specific groups exploring standardisation potential with respect to learning agents, these agents could benefit from many of the standards discussed under the deliberative component. Additionally there has been several related initiatives that could be considered for the realisation of the learning element.
For instance, the W3C Ontology-Lexicon Community Group\footnote{\url{https://www.w3.org/community/ontolex/}} has developed a lexicon model for ontologies\footnote{\url{https://www.w3.org/2016/05/ontolex/}} that can be used to enrich ontologies with linguistic information. While, the W3C Web Machine Learning Working Group\footnote{\url{https://www.w3.org/2021/04/web-machine-learning-charter.html}} aims to develop Web APIs that enable machine learning in the browser.

From an ontological perspective, both \citet{wong2009learning} and \citet{puerto2012automatic} demonstrate how various ontology learning techniques can be used to enhance manually crafted ontologies.  
While, \citet{merkle2019cooperative} show how reinforcement learning can be used to enhance the policies or strategies used by agents to complete their tasks.
That being said, it's worth noting that according to \citet{albrecht2018autonomous} many learning techniques are computationally complex making them unsuitable for many real world use case scenarios.  
When it comes to the agent learning semantic web models and frameworks, \citet{leite2013using} and \citet{leite2014case} propose high level ontology-driven hybrid agent architectures that include separate problem generator, critic and learning components that are used by the deliberative component in order to improve both the deliberative knowledge base and the reactive rules. 
While, \citet{young2016towards} demonstrate how spatial information about unknown objects together with their semantic web meaning can be used by robots to classify the unknown object. 
\citet{ghanadbashi2021using} in turn introduce their automatic goal generation model and a corresponding workflow that enables agents to evolve existing goals or create new goals based on emerging requirements. 
More broadly, \citet{asim2018survey} highlight that ontology learning has benefited from a variety of domains, namely natural language processing, machine learning, information retrieval, data mining and knowledge representation. The authors perform a comprehensive survey of existing work,  categorising them as linguistic,  statistic, and logic based.

\paragraph{Problem Generator \& Critic}

According to \citet{russel2013artificial} the problem generator suggests actions that will lead to learning in the form of new knowledge and experiences.  While the critic provides feedback to the agent in the form of a reward or a penalty.  Although the problem generator and the critic could vary greatly from an internal implementation perspective, there is a need for standardised vocabularies and APIs that can be used to manage synchronisation and communication between the various internal and external components.

From a vocabularies perspective, there has been some relevant work in terms of robotics, whereby \citet{buoncompagni2017ros} and \citet{kootbally2017overview} demonstrate how the Planning Domain Definition Language (PDDL) \footnote{\url{https://helios.hud.ac.uk/scommv/IPC-14/repository/kovacs-pddl-3.1-2011.pdf}} problem generator can make use of OWL reasoners to check for and solve issues with respect to norm compliance. 
As for models and frameworks, \citet{leite2013using} and \citet{leite2014case} present architectures whereby the agents perceive the effects that their actions have on the environment,  pass this information to the critic, which in turn informs the learning component about poor performance.  The learning component also recommends improvements and the problem generator is responsible for proposing new actions based on these recommendations. 
\citet{van2015semantic} focus on the weaker notion of norm compliance and propose a semantic framework that demonstrates how agents identify problems and adapt their behaviour in order to avoid violating norms.

\begin{table*}[!t]
\scriptsize
\caption{Intelligent software web agents controller component.}
{\color{black}\begin{tabular} 
{p{3.2cm} p{4.9cm} p{8.9cm}}
\toprule
&
\textbf{Standards} &
\textbf{Community Activities} 
\\
\bottomrule  
\rowcolor{gray!50} 
\multicolumn{3}{l}{\textbf{Perception \& Action Interpreters}} \\
\rowcolor{white}
Languages,  Ontologies \& \newline Vocabularies  &  FIPA\_Contract\_Net & messages \& message sequences \citep{challenger2018development}, generic planning ontology \citep{pham2011practical},  \\
\rowcolor{gray!25}
Models \& Frameworks  &  & ontological representation \citep{leite2013using, leite2014case},  semantic descriptions for unknown objects \citep{young2016towards} \\
\hline
\rowcolor{gray!50}
\multicolumn{3}{l}{\textbf{Execution Engine}}  \\
\rowcolor{white}
Languages,  Ontologies \& \newline Vocabularies  & OWL,  OWL-S, ODRL & control via policies \citep{tonti2003semantic},  Protune policy engine \citep{bonatti2010reactive}, normative language ontology \citep{fornara2018using, fornara2019using}, agent modelling language \citep{challenger2018development}, legal ontology \citet{palmirani2018pronto}  \\
\rowcolor{gray!25}
Models \& Frameworks  & LDP &  KAoS, Rei, Ponder comparison \citep{tonti2003semantic},  context broker architecture \citep{chen2004intelligent}, agent platform comparison \citep{pal2020review},  JACK \& OWL-S \citep{challenger2018development}, abstract state machines, Linked Data-Fu \& LDP \citep{kafer2018rule},  RDF/RDFS \& RuleML \citep{boley2001design},  Jason interpreter \citep{demarchi2018integration}, \\
\bottomrule   
\end{tabular}}
\normalsize
\label{tab:controller-component}
\end{table*} 

\subsection{Controller Component}
\label{sec:isa-controller}

The \texttt{Controller Component} is responsible for interpreting perceptions from sensors via the \texttt{Perceptions Interpreter}, devising execution plans that leverage the reactive and deliberative components, and executing the plans via the \texttt{Execution Plan Management}. In addition, this component is responsible for advising the actuators what action(s) need to be taken via the \texttt{Actions and Solutions Interpreter}.  
Although the tools and techniques that are needed to support agent control have not yet matured in terms of W3C standardisation efforts,  in \emph{Table}~\ref{tab:controller-component} we summarise preliminary research that could form the basis of initial standardisation discussions.  

\bigskip

\paragraph{Perception \& Action Interpreters} 

Irrespective of whether we are dealing with a web service or a web application, there is a need to define interfaces that can be used to interact with the agent. The perception interpreter is responsible for forwarding perceptions to the \texttt{Execution Engine}, while the action interpreter in turn is responsible for initiating actions forwarded by the \texttt{Execution Engine}.
When it comes to simple read and write operations, the Linked Data Platform (LDP)\footnote{\url{https://www.w3.org/TR/ldp/}} specification provides a set of best practices for an architecture that supports accessing, updating, creating and deleting Linked Data resources. 
However, said architecture would need to be amended to provide support for additional functions required in order to cater for interaction between intelligent agents.

\citet{challenger2018development} discuss the role played by messages and message sequences when it comes to agent interaction and highlight that they could be based on some standard such as FIPA\_Contract\_Net. More broadly, there are a range of FIPA standards\footnote{\url{http://www.fipa.org/specs/fipa00025/XC00025E.html},  \url{http://www.fipa.org/repository/ips.php3}} that could potentially be leveraged by intelligent software web agents.  From a goal encoding perspective, \citet{pham2011practical} propose an ontology than can be used for modelling planning problems that could be worked on by goal driven agents. In the frameworks proposed by \citet{leite2013using} and \citet{leite2014case} perceptions and actions are represented as ontologies,  however the authors focus on the general framework as opposed to the languages, vocabularies and ontologies that could be used for modelling perceptions and actions. \citet{young2016towards} in turn focus on leveraging external knowledge bases in order to obtain semantic descriptions for unknown objects. 

\paragraph{Execution Engine} 

The  \texttt{Execution Engine} is responsible for routing conditions to the \texttt{Reactive Component} and goals to the  \texttt{Deliberative Component}, thus the component needs to be able to handle both real-time and delayed responses.  Although there is a lack of specific W3C standardisation activity concerning the execution engine,  the semantic web community have proposed several tools and technologies that make use of web standards.  

\citet{boley2001design} discuss how RDF/RDFS and RuleML can together be used to develop simple reactive software agents.
\citet{tonti2003semantic} investigate how policies can be used to control agent behaviour by separating a systems functional and governance aspects. The authors compare and contrast the policy management approaches of the KAoS \citep{Uszok2003, uszok2004kaos}, Rei \citep{kagal2002rei, kagal2004authorization} and Ponder \citep{damianou2000ponder} policy languages and frameworks when it comes to controlling communication.  \citet{chen2004intelligent} also focus on policy enforcement,  proposing a context broker architecture that can be used to control the sharing and use of personal data.  When it comes to general policy languages, the Protune policy engined proposed by \citet{bonatti2010reactive} has also been used to control reactive behaviour.
\citet{fornara2018using} and \citet{fornara2019using} in turn propose a normative language ontology, derived from the ODRL standard, that could be used to control agent behaviour. While, \citet{palmirani2018pronto} introduce their legal ontology that could be used to design privacy preserving intelligent agents.
\citet{poulovassilis2006event} propose an abstract architecture that could guide the development of an event-condition-action reactive agent.  Whereas, \citet{kafer2018rule} use abstract state machines in order to model the internals of their reflexive agents and demonstrate the effectiveness of their proposal using Linked Data-Fu\footnote{\url{https://linked-data-fu.github.io/}} together with their Linked Data Platform implementation\footnote{\url{https://github.com/kaefer3000/ldbbc/}}.
\citet{challenger2018development} propose a platform independent multi-agent system development methodology and demonstrate its effectiveness using the JACK multi-agent system development framework together with OWL-S models.  Whereas, \citet{demarchi2018integration} demonstrate how the Jason interpreter can be adapted to benefit from ontological knowledge.  More broadly,  \citet{pal2020review} provide a comprehensive review of platforms that can be used to develop agent based systems, however their suitability for developing intelligent web agents is still and open area of research.
%

%% file: content/iswavision.tex
% !TEX root = ../main.tex
\section{Intelligent Software Web Agents: The Future}\label{sec:broaderperspective}

The goal of this section is to use insights gained from the analysis of the intelligent software web agent requirements and the hybrid agent architecture components,  in order to highlight existing research opportunities and challenges.  In addition, we take a broader perspective of the research by discussing the potential for intelligent software web agent as an enabling technology for emerging domains, such as digital assistants, cloud computing, and the internet of things.

\begin{table*}[!t]
\scriptsize
\rowcolors{2}{gray!25}{white}
\caption{Intelligent software web agents requirements assessment.}
\begin{tabular} 
{p{2.0cm} p{5.0cm} p{2.5cm} p{3.4cm} p{3.4cm} }
\toprule
\textbf{Functions} &
\textbf{Use Case Requirements} &
\textbf{Architectural \newline Impact} &
\textbf{\added{Opportunities}} &
\textbf{\added{Challenges}} \\
\bottomrule  
\rowcolor{gray!50}
\multicolumn{5}{l}{\textbf{Basic Functions}} \\
Autonomy  & consult relevant sources, devise an optimal schedule   & cross cutting & \added{ontology learning and reinforcement learning techniques} & \added{adopt a learning theory perspective} \\
Reactivity  & immediate response where possible & core & \added{event condition action rule languages and frameworks} & \added{reference architecture}   \\
%-
Pro-activeness  & scheduling goal,  explore alternatives  & core & \added{well established standards, techniques for representing \& reasoning over roles, behaviours, norms, beliefs, goals  \& plans} & \added{reference architecture} \\
Social ability  &  humans and agents & cross cutting & \added{policy \& norm languages} & \added{virtual organisations management techniques, policy \& norm standards}  \\
\hline
\rowcolor{gray!50}
\multicolumn{5}{l}{\textbf{Behavioural Functions}} \\
Benevolence & well meaning by design,  manage conflicting goals  & cross cutting& - & \added{benevolent by design}    \\
Rationality & rational by design & cross  cutting & -  &  \added{rational by design}  \\
Responsibility & mange access to information,  finds optimal schedule given a set of constraints  & task specific & \added{task environment requirements assessment} & \added{requirements elicitation techniques}  \\
Mobility & interacts with several other agents  & cross cutting & - & \added{discovery of services in mobile and ubiquitous environments, technological advances supporting agent mobility}  \\
\hline
\rowcolor{gray!50}
\multicolumn{5}{l}{\textbf{Collaborative Functions}}\\
Interoperability  & agreed/common schema &  core  & \added{established standards}  & \added{reference architecture}   \\
Communication &  push and pull requests &  core & \added{established standards,  norms \& policies} & \added{reference architecture}  \\
Brokering \newline services  &  collects information from a variety of sources   &  cross cutting & \added{established standards, semantic web service discovery \& composition techniques} &  \added{reference architecture}   \\
Inter-agent \newline co-ordination  &  agents support each other via information sharing  motivating & cross cutting & \added{policies \& norms} & \added{virtual organisations management techniques, policy \& norm standards}   \\\hline
\rowcolor{gray!50}
\multicolumn{5}{l}{\textbf{Code of Conduct Functions}} \\
Identification  & handle public and private information, may need to prove who they represent &  cross cutting & \added{unique identifiers \& authentication mechanisms} &  \added{agent architectures adaptation}   \\ 
Security  & protect against unauthorised access,  inappropriate use,  and denial of service & cross cutting & \added{access control, encryption} &  \added{agent architectures adaptation}   \\ 
Privacy  & handle personal information appropriately  & cross cutting & \added{privacy policies, anonymisation} &  \added{agent architectures adaptation}   \\ 
Trust  & manages information and scheduling accuracy, consults reliable sources  &  cross cutting & \added{trust frameworks and architectures} &  \added{agent architectures adaptation}   \\ 
Ethics & do no harm by design  & cross cutting & \added{legal policies, norms, general guidelines} &  \added{agent architectures adaptation}   \\ 
\hline
\rowcolor{gray!50}
\multicolumn{5}{l}{\textbf{Robustness Functions}}  \\
Stability  & available, reliable \& secure  &  robustness  & - & \added{attacker models}  \\
Performance & real time access to information, timely goal completion &  robustness & - &  \added{agent benchmarking tools}    \\
Scalability  & handles increasing requests,  data, \& task complexity & robustness & - &  \added{agent benchmarking tools}   \\
Verification  & checks information is correct, the reasoning is explainable & task specific \& \newline robustness &  \added{task environment requirements assessment, formal method, simulation}  & \added{requirements elicitation techniques, test driven development}  \\
\bottomrule   
\end{tabular}
\normalsize
\label{tab:architecture-future}
\end{table*}

\subsection{Opportunities and Challenges}

A condensed overview of the intelligent software web agents requirements analysis (focusing on the scheduling agent, which is the most complicated out of the three agents we examined), their impact from an architectural perspective, and the corresponding opportunities and challenges discussed below is presented in \emph{Table} \ref{tab:architecture-future}. 

\bigskip
\paragraph{Core aspects of the hybrid agent architecture}

The reactivity, pro-activeness, interoperability, and communication requirements are classified as core modules that are inherent to the hybrid agent architecture presented in the previous section.  When it comes to instantiating the architecture there are several standards and technologies that could be leveraged in order to realise the \texttt{Interface}, \texttt{Reactive}, and \texttt{Deliberative Components}.  Additionally, over the years the research community have proposed various approaches for semantic web service discovery and composition methods; event condition action rule languages and frameworks; and approaches for representing and reasoning over roles, behaviours, norms, beliefs, goals and plans, that could serve as a basis for developing a simple scheduling agent prototype. However, the suitability of the various proposals from both a practical perspective and a performance and a scalability perspective has yet to be determined.

Other challenges relate to the development of the \texttt{Controller Component}, which is responsible for internal co-ordination.  Existing proposals have focused on defining messages and message protocols; proposing ontologies for specifying norms and legal requirements; and controlling agent behaviour via policies.  Unfortunately, much of the work has focused on basic technology research, and many of the proposals have not been validated via prototyping. Here a reference architecture \cite{muller2008reference} could serve to bridge the gap between theory and practice and to identify potential open challenges that still need to be addressed from an architecture perspective.  

Additionally, the \texttt{Learning Component}, which is often discussed in the context of learning agents or normative/policy agents, has received little attention to date.  Broadly speaking, existing proposals focus on using ontology or reinforcement learning techniques to enhance the agents knowledge base,  or demonstrating how agents can adapt their behaviour based on changes in the environment. Here again, there is the need to determine the effectiveness of existing proposal in the form of a prototype.  When it comes to multi-agent learning in general, there are several survey articles (cf., \cite{panait2005cooperative, shoham2003multi, bloembergen2015evolutionary}) that could serve as the basis for the development of this component. While, from a practical implementation perspective, further research is needed to better understand its role in the overall architecture and what are the concrete standardisation needs.  

\paragraph{Task specific considerations}

Both the responsibility and verification requirements has been classified as task specific. In our motivating use case scenario, we identified three different types of agents, namely information agents, booking agents, and scheduling agents. Clearly this is not an exhaustive list of agent types, however considering the original semantic web vision has not yet been realised it is beneficial to revisit this simple use case, before moving on to more complex scenarios that can leverage intelligent web agents.
The key take home from an architectural perspective, is that we should strive to develop optional task modules that could serve a variety of use case scenarios. Here again there is an need to consider how such task specific modules could be integrated into a reference architecture and described in detail from a system design perspective. 
The agent task environment requirements assessment framework proposed herein can be used not only to perform a detailed analysis of various agent based use case scenarios,  but also to better understand the potential solutions and the technological and standardisations gaps that still exist.
Interesting directions for future work include adopting software engineering requirements elicitation techniques \cite{zowghi2005requirements, sommerville1998viewpoints, goguen1993techniques} and lessons learned \cite{davis2006effectiveness} in order to better understand the various use case requirements.

\paragraph{Cross cutting generic considerations}

The behavioural functions (i.e., benevolence, rationality, and mobility), code of conduct functions (i.e., identification, security, privacy, trust, and ethics) and two of the basic functions (i.e., autonomy and social ability) have been classified as cross cutting, as they need to be considered when it comes to the architecture as a whole and also the individual components.  Ideally they should be integrated into a reference architecture in the form of optional generic modules that can be used by the agent depending on the task that needs to be carried out.  Following common engineering practices, these modules would need to be described from a system design perspective.

In the early days of semantic web research, the behavioural functions (i.e., benevolence, rationality, and mobility) received some interest from intelligent software web agent researchers. In particular, researchers highlighted the need for agents to be benevolent and rational by design, to be capable of balancing self interest and group interests, to take on various roles and responsibilities, and to the need to cater for mobility from a robustness perspective.  However, when it comes to the proposed tools, technologies, and standards benevolence, responsibility,  and mobility requirements were not even mentioned in the corresponding papers.  
The lack of recent research in terms of intelligent software web agents behavioural functions is indicative of the communities diversification of interests and the need to better understand the needs of intelligent software web agents both from semantics and a deployment perspective, as argued by \citet{bernstein2016new}. 
The analysis of the intelligent software web agent requirements, and the standards, tools and technologies presented here-in is a first step towards better understanding the status quo and the requirements that should guide agents that leverage semantic web technologies.

When it comes to the code of conduct functions, there is a body of work from the semantic web community that has not been applied directly to the semantic web agent use case, that could potentially be leveraged in order to realise the proposed architecture.  In the following, we identify several interesting works that could potentially inform the design of our intelligent software web agents.
Broadly speaking, existing work in terms of identification focuses on access control for RDF \cite{Reddivari2005, Jain2006,Abel2007,Flouris2010,Dietzold2006, Gabillon2010,Kirrane2013} or demonstrating how policy languages can be used for the specification and enforcement of access restrictions \cite{Uszok2003,Kagal2003,Bonatti2009}.
Besides access control, security based research has primarily focused on applying encryption algorithms \cite{giereth2005partial, gerbracht2008possibilities, kasten2013towards, fernandez2017self} and digital signatures \cite{Kasten2014} to RDF data.
Work on privacy primarily focuses on applying and extending existing anonymisation techniques such that they work with graph data \cite{radulovic2015towards, HeitmannEtAl2017, lin2016isomorphism, Silva2017} or catering for the specification and enforcement of privacy preferences \cite{bonatti2019big, Sacco2011a}. 
When it comes to trust, \citet{artz2007survey} conducted a survey of existing trust mechanisms in computer science in general, and the Semantic Web in particular.  In addition, several authors have proposed trust frameworks and architectures \cite{ding2003trust,ding2005homeland, Schwabe2017, bizer2004using}.
When it comes to ethics, \citet{gordon2009rules} focus on requirements that are necessary for modelling and reasoning over legal rules and regulations,  whereas  \citet{palmirani2011legalruleml} extend RuleML in the form of LegalRuleML such that it can be used to model and reason over both legal norms and business rules.  
More generally, existing work at the intersection of intelligent agents and ethics \cite{dowling2000intelligent, dignum2018ethics} or behavioural aspects of intelligent agents \cite{wooldridge1995intelligent, wooldridge2001intelligent} could provide insights into the detailed design of these cross cutting generic modules.
Interestingly, the \emph{Ethics Guidelines for Trustworthy AI}~\cite{aihleg}, recently released by the European Commission only briefly mentions agent technologies, instead focusing on artificial intelligence in general. Thus, while this document serves as a useful starting point with respect to codes of conduct for agents, further work is needed to make these guidelines actionable from an intelligent agents perspective.

Basic agent functional requirements relating to autonomy and social ability serve as motivation for research concerning the role of policies and norms when it comes to controlling intelligent software web agent behaviour.  However,  there has been limited research by the semantic web community in terms of developing truly autonomous agents that are capable of interacting with other agents and the tools, technologies, and standards needed to enable agents to form virtual organisations in order to collaboratively solve problems.  Beyond the semantic web community \citet{van2006influence} propose gradual levels of autonomy that can be catered for via commitments and contracts. 
More generally, the technical opportunities and challenges relating to the field of agent based computing, identified by \citet{luck2005agent}, are also relevant from an intelligent software web agents perspective.  Primary considerations include: viewing autonomy from a learning theory perspective and examining social ability in terms of virtual organisations.  The authors also highlight the need for the advancement of tools and technologies to support scalable service discovery and composition, and semantic integration and additional research in terms of transparency, trust, reputation, and negotiation. 

\paragraph{Robustness considerations}

All four robustness functions (i.e., stability, performance, scalability, and verification) have simply been classified as robustness from an architectural perspective. These requirements need to be considered both when it comes to the detailed design of the system and the choice of technologies.  In the proposed architecture the \texttt{Performance Assessment} entity which is part of the \texttt{Task Environment} is responsible for evaluating the effectiveness of the system from both a functional and a non-functional perspective.  

Several of the requirements papers identified the need to evaluate existing proposals in terms of stability, performance and scalability. However, when it comes to the development of tools, technologies, and standards that could be used to evaluate the effectiveness of existing proposals, researchers have primarily focused on developing proof of concepts in the form of basic simulations or assessing formal aspects such as correctness, safety, and compliance.  Here again, we see evidence that intelligent software web agent research is more foundational than applied.  Considering the crucial role played by both functional and non functional testing from an engineering perspective, there is a need to develop testing strategies and benchmarks in order to advance the research further.
More broadly, when it comes to measuring robustness,  besides an array of individual performance evaluations for various query and reasoning engines, there is a body of work in relation to benchmarking that could be used/extended in order to benchmark the proposed architecture.  For instance, there are well established benchmarks, such as the Lehigh University Benchmark (LUBM)\cite{guo2005lubm} or the Berlin SPARQL Benchmark (BSBM) \cite{bizer2009berlin} and promising newcomers, such as the Linked Data Benchmark Council (LDBC) Social Network Benchmark~\cite{angles2020ldbc}. In addition, it may be possible to leverage existing benchmarking frameworks, such as the general entity annotator benchmarking framework (GERBIL)~\cite{usbeck2015gerbil} or the holistic benchmarking for big linked data framework (HOBBIT)~\cite{ngomo2016hobbit}.

\subsection{Semantic Web Agents as an Enabling Technology}

Moving beyond the original intelligent software web agent motivating scenario, the tools, technologies, and standards discussed herein could potentially have a much broader impact.  For instance, according to \citet{luck2005agent}, the semantic web community provides a semantically rich data model, vocabularies, and ontologies that can be used to describe media and services in a manner that facilitates discovery and composition; and allows for agent to agent information exchange.  In the following, we move beyond the original motivating scenario by highlighting the potential impact of intelligent software web agents on emerging domains, such as digital assistants, cloud computing, and the internet of things. 

\paragraph{Digital Assistants} Although well known voice assistants, such as Siri,  Alexa, and Cortana, are not as sophisticated as the Knowledge Navigator concept proposed by  Sculley \citep{sculley1987odyssey} (when he was the chief executive officer at Apple) the technology has been embedded in various smart home and smart phone products. Common features include sending and receiving text messages and emails, making calls, setting timers and reminders, and control of hardware (e.g., thermostats, lights, audio, video) \citep{hoy2018alexa}. 
However, in order to realise Sculley's Knowledge Navigator these voice assistants need to be enhanced with data discovery and reasoning capabilities, which are at the core of envisaged intelligent software web agents.
Considering the sensitive personal nature of the data often captured by such agents,  intelligent software web agents could also be employed in order to provide users with more control and transparency with respect to personal data processing.

\paragraph{Cloud Computing} Cloud computing has been around for quite some time, however as technology rapidly evolves so too does the service offering, for instance edge computing is a paradigm whereby computation is performed closer to where the data is consumed \citep{shi2016edge}. New data infrastructure initiatives,  such as GAIA-X \citep{braud2021road}, envisage virtual data spaces developed on top of federated infrastructure (including high performance computing and edge systems), where data sovereignty and secure exchange are built-in by design.  Recently, the term \emph{private 5G networks} is used to refer to industrial networks that require increased reliability, low latency, and strong security \citep{hoy2018alexa}.   
In this context, intelligent software web agents could potentially play a major role both from a resource allocation and a governance perspective.  In the case of the former, agents could take on a coordinating role when it comes to virtual organisation / private network formation and monitoring.  In the case of the latter, both data and service providers could encode usage constraints and provenance trails using policy languages and ontologies in a manner that supports agent based negotiation and automated compliance checking.

\paragraph{The Internet of Things} The W3C Web of Things initiative\footnote{\url{https://www.w3.org/WoT/}} focuses on building on existing Web standards in order to facilitate data integration across various IoT platforms.  Here, semantic technologies have already been used in order to describe things\footnote{\url{https://www.w3.org/TR/wot-thing-description/}} and facilitate thing discovery\footnote{\url{https://www.w3.org/TR/wot-discovery/}}.  
When it comes to the intersection of intelligent software web agents and the internet of things,  semantic web agents could also play a crucial role in terms of coordinating the usage, management, and governance of things. Additionally, the standards, tools, and technologies discussed herein could provide support for analytics needed in order to optimise supply and value chains that make use of IoT technologies.

%% file: content/iswaconclusion.tex
% !TEX root = ../main.tex

\section{Conclusions}\label{sec:conclusion}

Motivated by the desire to further advance existing research into intelligent software web agents, in this paper we revisited the original use case scenario proposed in the seminal semantic web paper from a gap analysis perspective.  We started by collating and summarising requirements and core architectural components relating to intelligent software agents in general.  Following on from this, we used the intelligent software agent requirements to both further elaborate on the semantic web agent motivating use case scenario, and to summarise and classify existing semantic web agent literature.  
We subsequently used the insights gained in order to propose a hybrid semantic web agent architecture that guided our discussion with respect to relevant standards, tools, and technologies.  Following on from this, we used the functional and non-functional agent requirements together with the scheduling agent use case requirements to better understand the opportunities and challenges concerning the realisation of intelligent software web agents. Finally, we broadened the discussion and highlighted the potential of intelligent software web agent as an enabling technology for digital assistants, cloud computing, and the internet of things. 

Key outputs include: (i) a task environment requirements assessment framework, based on agent requirements gleaned from the literature,  that could be used to perform an in-depth assessment of various agent use case scenarios; and (ii) a hybrid architecture and the corresponding assessment of existing standards, tools, and technologies, which serves as the basis for developing a reference architecture that can be used to realise the original intelligent software web agent vision and to build the foundations needed in order to support more complex use case scenarios. 

Based on our analysis, there are a number of gaps that still need to be addressed in order to move the intelligent software web agent vision forward.
Firstly, from an architectural perspective, there is a need to develop a reference architecture that could serve to bridge the gap between theory and practice, and to identify potential open research challenges that still need to be addressed.  
Secondly, from an implementation perspective there is a need to better understand the specific requirements relating to the cross cutting behavioural functions (i.e., benevolence, rationality, and mobility), code of conduct functions (i.e., identification, security, privacy, trust, and ethics), and basic functions (i.e., autonomy, and social ability),  the adaptations/extensions needed to existing tools and technologies, and insights into how these tools and technologies fit together with core intelligent software agent technologies and with each other.
Finally, from a robustness perspective, there is a need to develop/extend existing benchmarks such that they can be used to both validate and assess the performance and scalability of various instantiations of our hybrid agent architecture.